\pdfoutput=1

\documentclass[11pt]{article}

\usepackage[]{acl}

\usepackage{times}
\usepackage{latexsym}
\usepackage{multirow}
\usepackage[T1]{fontenc}

\usepackage{titlesec} 
\titleformat{\section}{\bfseries\large}{\thesection}{0.23cm}{}%

\usepackage{array}
\makeatletter
\newcolumntype{V}[1]{>{\topsep=0pt\@minipagetrue}p{#1}<{\vspace{-\baselineskip}}}
\makeatother

\usepackage[utf8]{inputenc}

\usepackage{microtype}
\usepackage{comment}
\usepackage{booktabs}
\usepackage{amsfonts}
\usepackage{amsmath}
\usepackage{graphicx}
\usepackage{inconsolata}
\usepackage{setspace}
\usepackage{lipsum}
\usepackage{bm}
\usepackage{booktabs}
\usepackage{enumitem}
\usepackage[export]{adjustbox}
\usepackage{caption}
\usepackage{subcaption}
\definecolor{dgreen}{HTML}{426137}
\definecolor{ired}{HTML}{B42F24}
\definecolor{eblue}{HTML}{2F71BA}
\newcommand{\teb}[1]{\textcolor{eblue}{#1}}
\newcommand{\tdg}[1]{\textcolor{dgreen}{#1}}
\newcommand{\tir}[1]{\textcolor{ired}{#1}}

\newcommand\env{{DescribeWorld}}
\newcommand{\ttt}[1]{\texttt{#1}}

\newcommand{\nattention}{{\mathrm{Attention}}}
\newcommand{\nsoftmax}{{\mathrm{Softmax}}}
\newcommand{\nmaskedmean}{{\mathrm{MaskedMean}}}
\newcommand{\nresblock}{{\mathrm{ResidualBlock}}}
\newcommand{\nargmax}{{\mathrm{Argmax}}}
\newcommand{\ntanh}{{\mathrm{Tanh}}}

\newcommand{\nlinear}{{\mathrm{Linear}}}

\newcommand{\ngru}{{\mathrm{GRU}}}

\title{One-Shot Learning from a Demonstration \\ with Hierarchical Latent Language}

\author{Nathaniel Weir$^\spadesuit$ \:\:\:\: Xingdi Yuan$^\dag$  \:\:\:\: Marc-Alexandre C\^ot\'{e}$^\dag$ \\ 
\textbf{Matthew Hausknecht$^{\dag}$  \:\:\:\: Romain Laroche$^{\dag}$ \:\:\:\: Ida Momennejad$^{\dag}$} \\ \textbf{Harm Van Seijen$^{\dag}$ \:\:\:\: Benjamin Van Durme$^{\spadesuit\ddag}$} \\
$^\spadesuit$Johns Hopkins University \:\:\:\: $^\dag$Microsoft Research
\:\:\:\: $^\ddag$Microsoft Semantic Machines\\
\ttt{nweir@jhu.edu} \:\:\:\: \ttt{eric.yuan@microsoft.com}
}

\begin{document}
\maketitle
\begin{abstract}
Humans have the capability, aided by the expressive compositionality of their language, to learn quickly by demonstration. 
They are able to describe unseen task-performing procedures and generalize their execution to other contexts. 
In this work, we introduce \env{}, an environment designed to test this sort of generalization skill in grounded agents, where tasks are linguistically and procedurally composed of elementary concepts. 
The agent observes a single task demonstration in a Minecraft-like grid world, and is then asked to carry out the same task in a new map.
To enable such a level of generalization, we propose a neural agent infused with hierarchical latent language---both at the level of task inference and subtask planning.
Our agent first generates a textual description of the demonstrated unseen task, then leverages this description to replicate it.
Through multiple evaluation scenarios and a suite of generalization tests,
we find that agents that perform 
text-based inference 
are better equipped for the challenge under a random split of tasks. 
\end{abstract}

\section{Introduction}
Humans are highly capable of learning by example. 
If a child watches their school teacher draw a purple winged elephant then recite the alphabet backwards, they can replicate the sequence of activities at home with relative ease.
This is in no small part due to the human ability to leverage the compositionality of language in order to comprehend new situations composed of familiar concepts~\cite{chomsky-1957-syntactic}. 
The child can restate the demonstration in words (as we did above), naturally decomposing it into its distinct subcomponents (the drawing, and the alphabet), which are themselves procedurally compositional (e.g., {\it ``pick up purple marker, \dots''}). 
Humans use their linguistic understanding of a task's hierarchical compositionality to generalize it to a new context; 
without this generalization, we might expect a child would overfit to the specifics of the classroom context.

In this work, we explore whether grounded artificial agents can similarly generalize from a demonstration: a single expert trajectory accomplishing a task. Specifically, we pose a setting where an agent observes a demonstration of a never-before-seen task, then must perform the task in a new context.

We construct \env{}, an environment containing a dataset of high-level tasks involving building recipes, navigation, and interaction with objects and terrains.\footnote{Examples available at \url{describeworld.github.io}; dataset and code will be released publicly.}
Test tasks are distinct from training tasks, but they are procedurally composed of 
the same subtasks and low-level actions.

\begin{figure}[t!]
    \centering
    \includegraphics[width=\columnwidth]{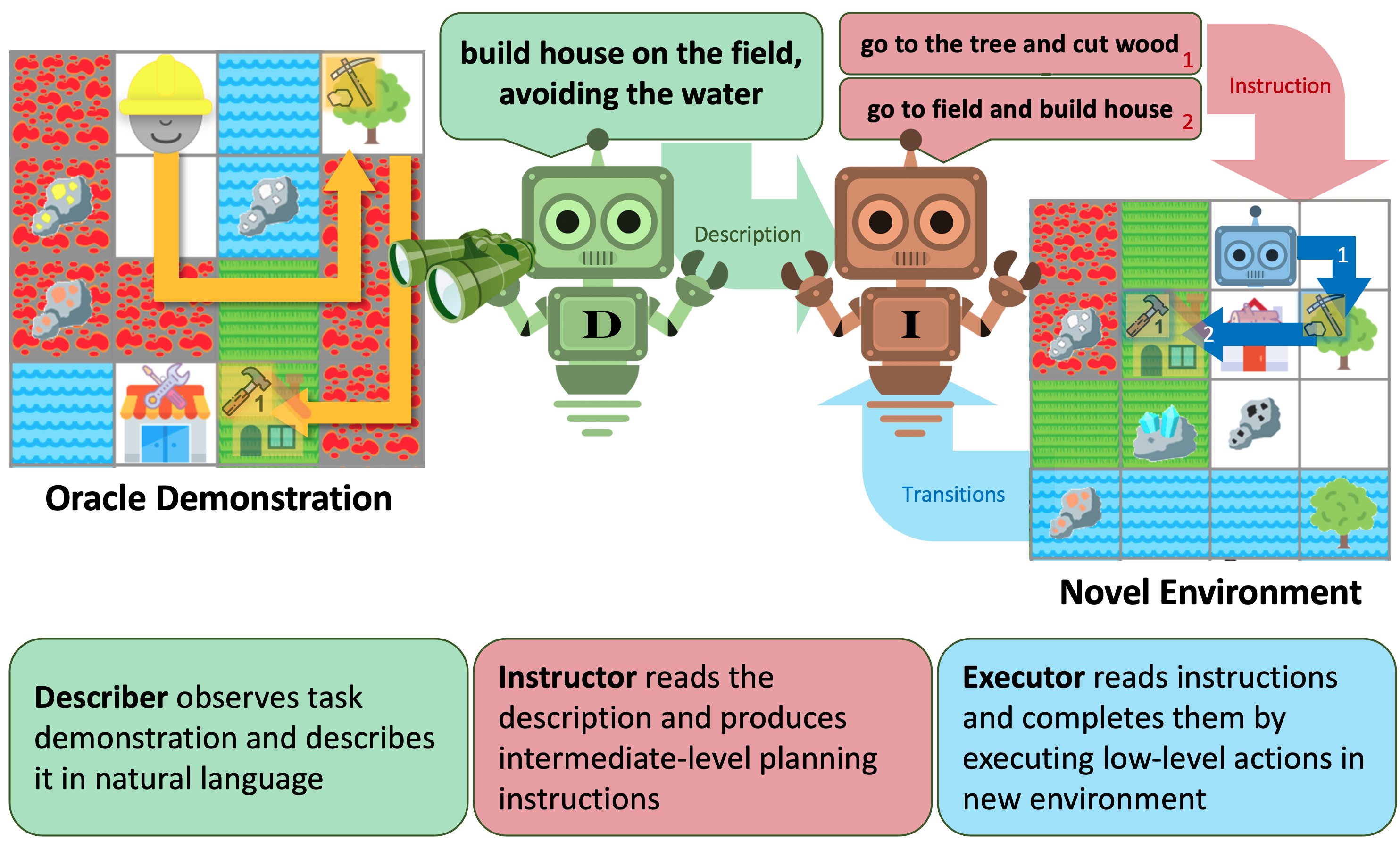}
    \caption{
    Framework for learning from demonstration via latent language. 
    The \tdg{\textbf{\underline{D}escriber} module} observes an oracle demonstration of an unseen task and describes it in text. Given the generated description, the \tir{\textbf{\underline{I}nstructor} module} infers necessary subtasks, accomplished by the \teb{\textbf{\underline{E}xecutor} module} via low-level control actions.
    }
    \label{fig:overview}
    \vspace{-1em}
\end{figure}

As humans leverage language to perform such generalization, we follow recent work~\citep{ruis-etal-2020-benchmark} by designing, alongside a traditional random task split, a suite of benchmark splits that require learning systematic rules governing how linguistic variation affects a task's subtask `recipe.'
For example, the agent might be trained to build a \ttt{pig barn} and an \ttt{iron shrine}, then during testing must build the unseen composition \ttt{pig shrine}. 

To perform in this task environment, we devise a novel three-level Hierarchical Latent Language Policy (HLLP) agent that represents both high-level tasks ({\it ``build a house on field''}) and subtask plans ({\it ``cut wood''}) in natural language. 
As depicted in \autoref{fig:overview}, this effectively recasts the challenge of learning from demonstrations as a) describing the demonstrated unseen task, then b) following the predicted description in a new map. 
The agent uses text representations at two levels of abstraction: identifying top-level verbalized tasks (via a \textit{describer} module), and identifying a sequence of intermediate-level subtasks (via \textit{instructor}). We train the agent via imitation learning on synthetic text associated with oracle actions.

Our novel testing scenario for \env{} is \textbf{demonstration following}, where the agent 
must replicate a demonstrated task in a new random map. 
Given its challenging nature, we also evaluate a simpler scenario, \textbf{description following}~\cite{weller-etal-2020-learning},
which assumes that the agent instead has access to a gold task description. 
This ablated variant allows us to examine performance at lower levels of abstraction by asking: were an agent to successfully describe an unseen task using NL, could it then follow the task in a new context?

We contrast approaches that leverage latent language policies versus those that instead use continuous representations.
We find that modeling agent policy as latent natural language improves the ability to generalize to demonstrations of unseen tasks.

\subsection{Contribution}
We frame the contribution of our new demonstration following environment
and our proposed 
HLLP agent
in terms of \citet{lake-murphy-2021-word}'s five desiderata for a computational theory of semantics characteristic of human language use:\\
\noindent\textbf{1. Describing, or understanding the description of, a perceptually present scenario:} the {HLLP} agent receives as input a multi-modal
demonstration of a task, and expresses it in text so as to generalize into a new randomly-generated map. \\
\noindent\textbf{2. Choosing words on the basis of internal desires, goals, or plans:} the agent uses natural language to both describe a demonstrated high-level task, as well as to verbalize intermediate-level subtasks to complete at the level of control policy.\\
\noindent\textbf{3. Responding to instructions and requests appropriately:} the agent iteratively executes action sequences against the task environment in order to follow the high-level descriptions and low-level instructions it produces for itself.\\
\noindent\textbf{4. Producing and understanding unseen conceptual combinations:} test demonstrations show unseen high-level tasks composed linguistically and procedurally of known concepts.\\
\noindent\textbf{5. Changing one’s beliefs about the world based on linguistic input:} demonstrations convey environmental constraints -- e.g. that walking on lava yields a penalty--- that the agent must verbalize and act upon via low-level control policy.

\section{Related Work}
\paragraph{Latent Language Policy Agents}
Natural language has been proposed as a medium for conveying task-specific goals~\cite{karch-etal-2020-language} and constraints~\cite{yang-etal-2021-safe} to grounded reinforcement learning agents.
\citet{andreas-etal-2018-learning} show the benefit of reparamatrizing a continuous policy search into discrete text space for various few-shot `learn-the-rule' tasks.
They suggest that such "latent language policy" (LLP) models are a promising avenue for generalization on the basis of language learning. More recent work has applied LLPs to real-time strategy games~\cite{hu-etal-2019-hierarchical, jacob-etal-2021-multitasking}, while \citet{chen-etal-2021-ask} show that LLPs trained to generate and follow crowdsourced instructions can perform few- or zero-shot simple crafting tasks in a small grid world.  
Ours is a similar style of environment, though our high-level tasks are more complex, extending beyond individual crafting recipes.\footnote{Performance by \citet{chen-etal-2021-ask}'s model degrades for crafting recipes with 5 `steps', while ours have upwards of 16.}
\citet{jiang-etal-2019-language} train hierarchical synthetic language policy agents to accomplish a set of shape-arranging tasks in a MuJoCo-based environment. They find that language can improve performance on a simple form of systematic generalization (holding out tasks where the first half of instructions include the word ``red''). 

\paragraph{Grounded Language Environments}
Several recent language grounding environments study an embodied agent given high-level task descriptions and/or instructions to follow, e.g., LANI~\cite{LANI}, Room2Room~\cite{Anderson2018VisionandLanguageNI}, ALFRED~\cite{ALFRED}. ALFRED has a similar notion to ours of task decomposition, where tasks and subtasks are expressible via NL instruction. However, due to limitations of their underlying 3D engine, they cannot evaluate complex crafting tasks as a means to target systematic generalization.
\citet{chevalier-etal-2018-babyai} and \citet{hill-etal-2019-environmental} investigate compositional rule learning for navigational and pick-up/put-down skills using a synthetic language of instructions in 2D and 3D environments, respectively. 
\citet{jiang-etal-2020-wordcraft} consider a text-based environment in which agents must infer zero-shot concept combination recipes using common sense.
\citet{ruis-etal-2020-benchmark} construct a grounded instruction following benchmark that evaluates many types of systematic generalization. 
Our effort builds upon theirs, introducing a novel scenario (demonstration following) as well as tasks with longer trajectories, subtask dependencies, and new action types (building/placing). 

\begin{figure}[t!]
    \centering
    \includegraphics[width=\columnwidth]{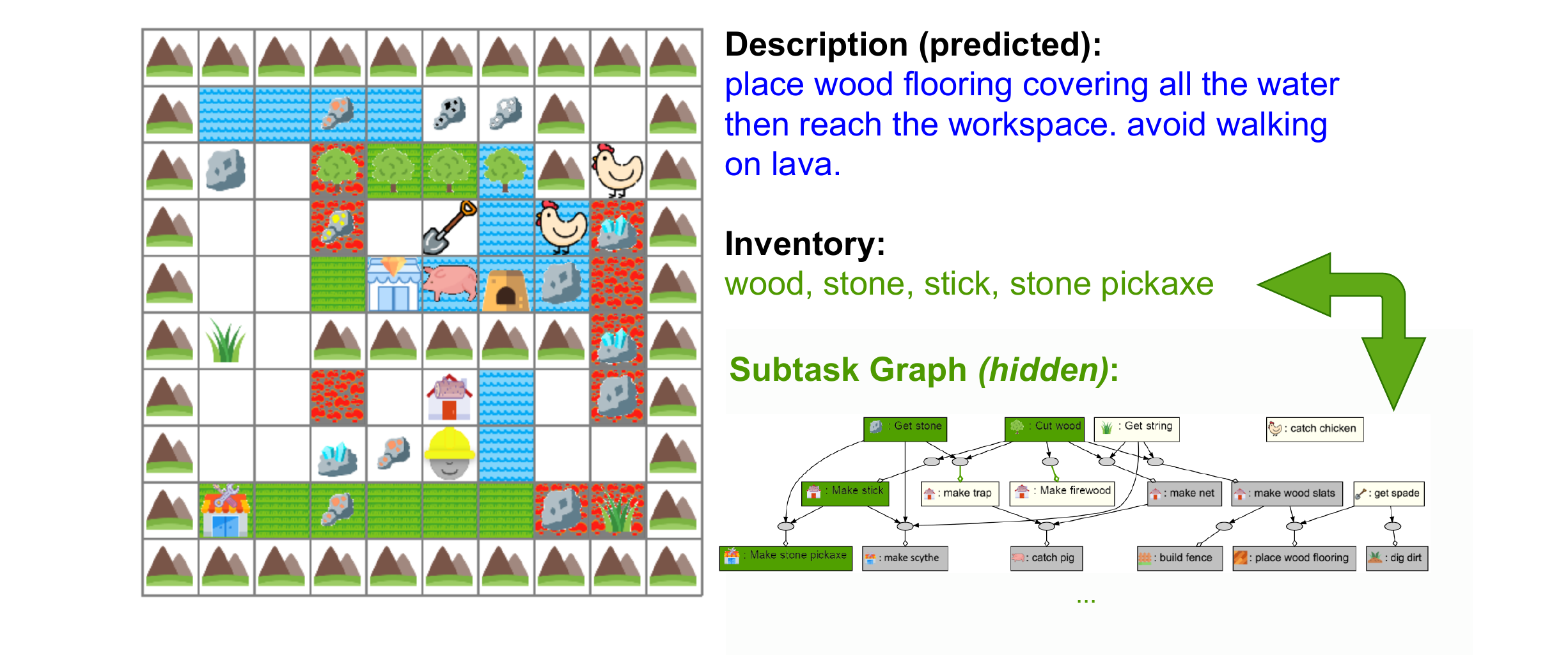}
    \vspace{2mm}
    {\footnotesize
    \begin{tabular}{lrlr}
    \toprule    
    \multicolumn{3}{l}{Unique Tasks (End Goals + Terr. Consts.)} &  $10604$\\
    \multicolumn{3}{l}{Unique End Goals}     &  $2651$  \\
    Objects & $29$ &
    Pickable Objects & $11$ \\
    Craftable Items & $19$ &
    Buildable Structures & $13$  \\
    Placeable Terrains & $7$ &
    Natural Terrains & $3$ \\
    \bottomrule
    \end{tabular}
    }
    \caption{\env{} overview.
    Maps are symbolic images,
    while the task description, predicted by the agent from a demonstration, and the inventory, reflecting subtask completion, are encoded in text.
    }
    \vspace{-1em}
    \label{fig:env_overview}
\end{figure}
\paragraph{Language-Based Generalization in Humans and Models}
\citet{lake-baroni-2018-generalization} show that RNN-based sequence models struggle to perform systematic compositional generalization based on abstract linguistic rules, while humans are extremely effective at it given few examples~\cite{lake-etal-2019-human}. 
Other recent NLP work explores training language models to perform few-shot task generalization given textual task descriptions~\cite{weller-etal-2020-learning,mishra-etal-2021-crosstask,wei-etal-2021-finetuned}.

\paragraph{Meta-Learning}
One way to achieve generalization is to learn strategies that can quickly adapt to novel tasks by leveraging past experiences \cite{schmidhuber:1987:srl,Thrun1998LearningTL,Bengio2007OnTO}. Specifically, our experimental setup falls under the zero- and few-shot imitation learning category \cite{Duan2017OneShotIL,Yu2018OneShotIF,Pan2020ZeroshotIL,Zhou2020WatchTL}, where our approach receives a \textbf{single} demonstration to solve novel tasks.

\section{\env{} Environment}
\label{sec:summary}
\env{} is a 2D grid world implemented atop the \textbf{Mining} domain from \citet{sohn-etal-2018-hierarchical}. 
The procedurally generated map (\autoref{fig:env_overview}) is an 8x8 grid (with surrounding walls); cells can contain terrains and objects. The agent can perform movement, \texttt{use}, and \texttt{place} actions in order to complete subtasks that either add resources to its inventory, build items, or place craftable terrains at the agent's location. Details can be found in \autoref{app:appendix-env} and on our \href{https://describeworld.github.io}{project webpage}. 
The set of possible subtasks and their dependencies (depicted in Appendix \autoref{fig:full_graph}) is constant across all tasks;
we combine subtasks in unseen ways to form unique high-level tasks
to be learned from demonstration. 
\begin{figure}[t!]
\small
    \centering
\includegraphics[width=.9\columnwidth]{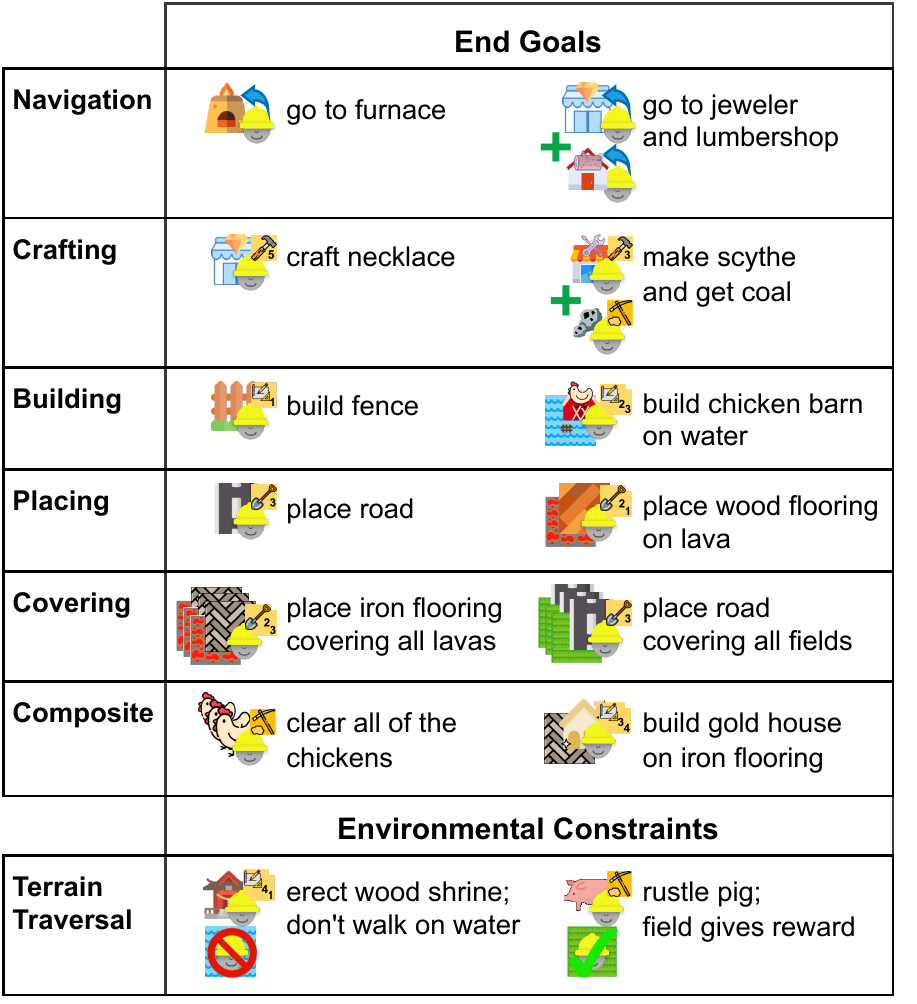}
    \caption{Categories of end goals and environmental constraints parametrizing high-level tasks.
    }
    \vspace{-0.5em}
    \label{fig:task-categories}
\end{figure}
\subsection{Compositional Tasks and Subtasks}
Tasks and subtasks in \env{} exhibit procedural and lexical compositionality. A list of high-level task categories is shown in \autoref{fig:task-categories}. Tasks may also be parameterized by \textit{environmental constraints}--namely, that traversing a particular type of terrain will produce either a reward or a penalty.

Certain building and placing subtasks optionally accept a special ingredient material, e.g. \texttt{gold} \texttt{house}. The recipes for these subtasks comprise those needed to acquire the material plus those needed to build the object. 
All \texttt{gold} items require smelted gold, while all \texttt{house}s, whether they are \ttt{silver}, \ttt{gold}, or regular, require \ttt{wood} \ttt{slats}, and \ttt{iron}. These subtasks require a pair of build-key actions to complete: the first uniquely determines the type of object to build, while the second determines which special material should be used. 
The action to specify a given special ingredient is constant across all special recipes.
Further details of such subtasks are shown in Appendix \autoref{tab:compositional_recipes}.

\subsection{State Representation}
The state at time step $t$ is represented as a tuple $(M_t, I_t)$, where map $M_t$ is a symbolic $8 \times 8 \times 3$ tensor with channels for agent, item, and terrain. Inventory $I_t$ is a text representation (comma separated) of the currently-held items, e.g. \ttt{wood,} \ttt{stone,} \ttt{spade}. 
There is a step penalty of $-1$, and we track the number of traversals over reward- and penalty-giving terrains; rewarding cells can only be triggered once per game.  Trajectories end upon end goal completion, or hitting a 300-step time limit.

\subsection{Oracle}

We implement an oracle that 
navigates the gridworld and completes high-level tasks. 
The oracle computes the set of all necessary subtasks required to complete the high-level task. 
It then computes the intersection of necessary and currently eligible (i.e. prerequisite-satisfied) subtasks, 
then chooses one to complete according to a canonical order.
This process is repeated until the high-level task is completed. Example trajectories are provided in Appendix \autoref{fig:oracle_unrolls}.
The oracle is used both to generate trajectories for demonstration following (rolling out a trajectory from start to finish), as well as to provide gold instructions and executions during imitation learning 
(i.e. used on-the-fly to generate the next step towards completing the next subtask).
In the former case, in order to convey environment-specific constraints such as rewards/penalties for stepping on particular terrain types, we ensure that it traverses all terrain types at least once. Ensuring traversal of all terrains can require a navigational detour of a couple steps.

\begin{figure}[t!]
    \centering
    \includegraphics[width=.9\columnwidth]{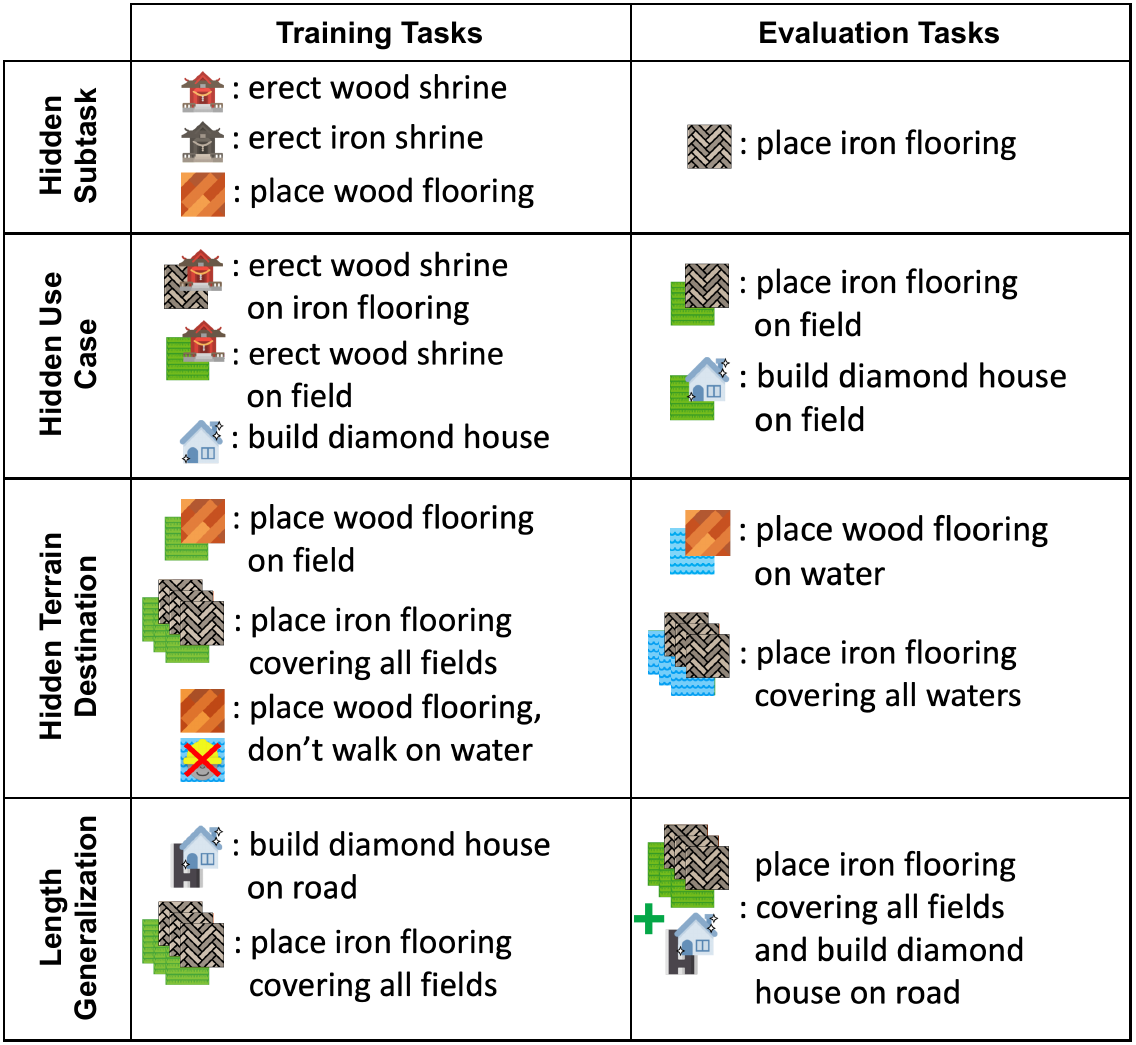}
    \caption{Data splits testing for systematic generalization in demonstration following agents}
    \vspace{-0.5em}
    \label{fig:splits}
\end{figure}
\subsection{Data Splits}
We introduce a suite of train/test splits, depicted in \autoref{fig:splits}, each of which requires a particular form of rule-based systematic generalization in demonstration following agents.

\noindent\textbf{Random Split}\quad
We compare against a simple random 70/30 split, where tasks are sorted by hashing the text of their end goal, ignoring terrain rewards/penalties. 
The random split test is nontrivially challenging due to complex subtask dependencies and unseen randomly-generated maps.
\begin{figure*}[t!]
    \centering
    \includegraphics[width=\textwidth]
    {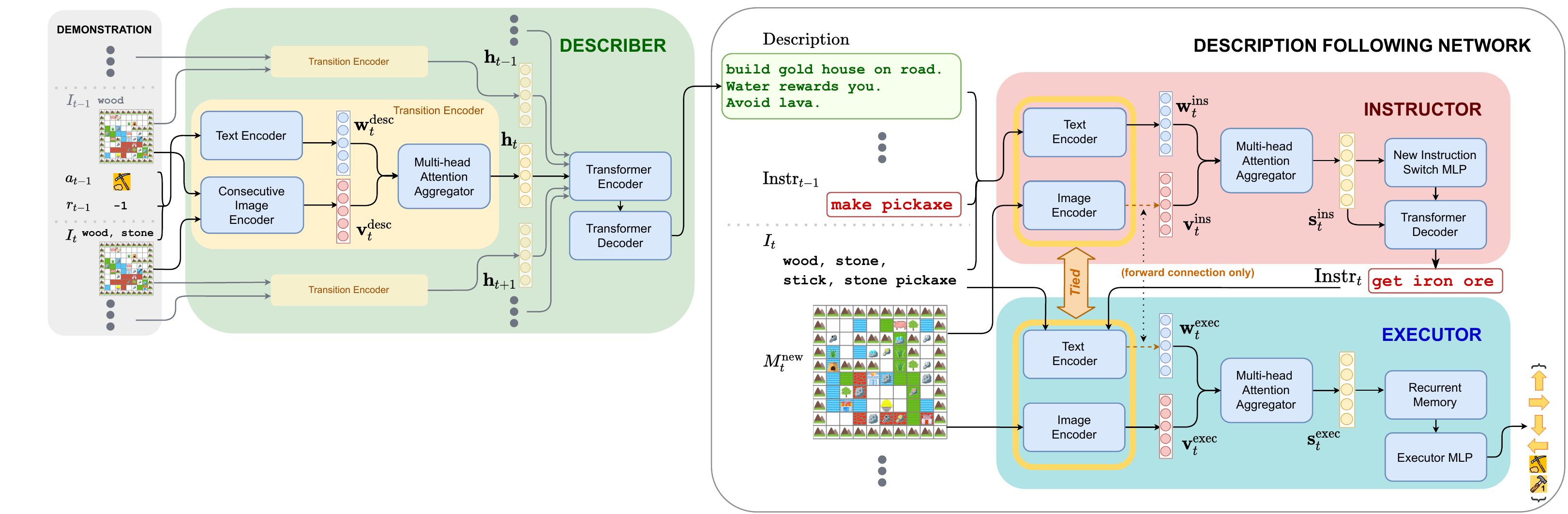}
    \caption{Architecture of hierarchical latent language policy agent. The \tdg{describer} module decodes a description of a demonstration in map $M^\text{dem}$, then 
    the 
    \tir{instructor}/\teb{executor} modules replicate the task in new map $M^\text{new}$.}
    \vspace{-0.5em}
    \label{fig:agent_2_arch}
\end{figure*}

\noindent\textbf{Hidden Subtask}\quad
This split requires procedural generalization on the basis of ingredient/object composition. We remove from the training data all end goals involving the subtask \texttt{place iron flooring}, but leave in all other tasks that involve other types of \texttt{flooring}, and those that use the \texttt{iron} special ingredient. 
We repeat the procedure with \texttt{erect pig shrine} and \texttt{build diamond house}. Appendix \autoref{tab:compositional_recipes} lists the building recipes for these subtasks, plus those left in the training set with which they linguistically overlap; those serve as the source of generalization. 
The test set contains all tasks 
that involve any of the three unseen subtasks.\footnote{We leave out tasks requiring \texttt{cover}ing terrain from the hidden subtask and use case test sets due to agents' low completion rate on the category under the random split. } This challenge is twofold: the agent must learn that modifiers like \ttt{pig} and \ttt{diamond} correspond to a required set of subtasks, plus a fixed specification action when building a structure.

\noindent\textbf{Hidden Use Case}\quad
This split requires generalization of a subtask learnt in one isolated use case. 
We remove from training all tasks involving \texttt{diamond house}, except for the plain task \texttt{build} \texttt{diamond} \texttt{house}.
At test time, the agent must use the subtask in all other end goals, e.g. \texttt{build} \texttt{diamond} \texttt{house} \texttt{on} \texttt{field}.
We repeat the process for \ttt{place} \ttt{road} and \ttt{make} \ttt{goldware}. 
We also test the generalization of \ttt{iron} \ttt{flooring} appearing during training only as a destination, e.g. in \ttt{build} \ttt{house} \ttt{on} \ttt{iron\ttt} \ttt{flooring}. The agent is tested on all other use cases, e.g. \ttt{place} \ttt{iron} \ttt{flooring} \ttt{on} \ttt{field}. 

\noindent\textbf{Hidden Terrain Destination}\quad
This split requires generalization of terrains as not only sources of traversal penalty/reward, but also as a building destination. We hold out all tasks that involve the terrain \texttt{water} as a destination, e.g. in \texttt{build house on water}.
We leave in tasks that use other terrain types, e.g. \texttt{lava} and \texttt{field}, as destinations. We also leave in tasks that involve \texttt{water} as a terrain constraint, as in \texttt{build house. don't walk on water}. This split therefore requires agents to generalize the fact that \texttt{water} can also serve as a destination from the dual roles of other terrains.

\noindent\textbf{Length Generalization}\quad
Neural sequence models show to fail to generalize to task lengths longer than those seen in training~\cite{
ruis-etal-2020-benchmark}. 
We test for this capacity by holding out tasks with the top 10\% longest oracle trajectories.

\section{\textls[-30]{Hierarchical Latent Language Policy Agent}}
\label{sec:model}
We design a three-layer hierarchical latent language policy (HLLP) agent to perform one-shot demonstration following. 
The \textbf{describer} module observes oracle demonstrations and describes them in text. The description following \textbf{instructor} and \textbf{executor} modules work in tandem to generate intermediate-level NL instructions and choose low-level actions. 
We train modules to use a compositional, canonical subset of English as a means for efficient policy communication with other modules.\footnote{This design choice is in contrast with existing work, e.g. \citet{hu-etal-2019-hierarchical, chen-etal-2021-ask}, that trains LLPs on crowdsourced NL instructions with high variation. We do not see the high variability of naturally occurring language as necessary for our agents to communicate policy decisions; the describer need not generate verbose linguistic alterations in order to effectively convey task-relevant information to other modules.}
We thus parametrize our agent's policy via text description $D$ and instruction sequence $\text{Instr}_{1} \dots \text{Instr}_i$.

\vspace{-3mm}
\resizebox{.95\columnwidth}{!}{
  \begin{minipage}{\columnwidth}
  \begin{align*}
D = f_{\text{descr}}(&M_{1:n}^{\text{dem}}, I_{1:n}^{\text{dem}}, a_{1:n}^\text{dem}, r_{1:n}^\text{dem})\\
\text{Instr}_i = f_{\text{instr}}(&M_i, I_i, \text{Instr}_{i-1}; D)  \\ 
a_i = f_{\text{exec}}(&M_{1:i}, I_{1:i}, a_{1:i-1};  \text{Instr}_{1:i}) 
\end{align*}
  \end{minipage}
}

\noindent\textbf{\textbf{Describer} module}\quad
\label{section:hllp:describer}
Depicted in (\autoref{fig:agent_2_arch}, left), this is a basic transformer-based ``video summarization'' model.
It takes a demonstration (i.e., sequence of transitions) as input. 
A transition at time step $t$ is a 5-tuple including the previous step's symbolic image $\text{M}_{t-1}$, the action taken $a_{t-1}$, the resulting reward $r_{t-1}$, the resulting symbolic image $\text{M}_{t}$, and the text enumerating the new inventory 
$\text{I}_{t}$. 

For each time step $t$, we use an image encoder to encode $M_{t-1}$ and $M_{t}$, and a text encoder to encode the concatenation of $a_{t-1}$, $r_{t-1}$, and $I_{t}$. 
The resulting encodings are aggregated using an attention mechanism into a single transition representation.
To obtain a single \textit{demonstration} representation, we use a second transformer encoder over the sequence of transition encodings
, then use a standard 
attention-equipped 
transformer decoder to generate a description of the demonstrated task.

\noindent\textbf{\textbf{Instructor} module}\quad
\label{section:hllp:instructor}
Our framework for generating and following instructions given a task description is similar to that of \citet{hu-etal-2019-hierarchical}, except we use a language model decoder instead of a classifier and compute separate state encodings for the two modules. 
 At each time step, the \textit{instructor} module (\autoref{fig:agent_2_arch}, upper right) computes a multimodal state representation via attention-based aggregation of separate encodings of the textual and image components of the state observation. The text representation is a transformer encoding of the task description concatenated with the inventory text, while the image representation is a convolutional encoding of the map.
The state representation is passed to the `new instruction' classifier, which determines whether to decode a new instruction or copy the previous timestep's.\footnote{This is necessary because of a lack of a state cue signifying the need for a new instruction, e.g. a change in inventory in \citet{chen-etal-2021-ask}.}

\noindent\textbf{\textbf{Executor} module}\quad
\label{section:hllp:executor}
This module (\autoref{fig:agent_2_arch}, lower right) computes a combined state representation using the same encoder parameters, but using the generated instruction text instead of the task description. The state representation is used to update a recurrent memory cell, the hidden state of which is fed to an MLP classifier over low-level actions.


\subsection{Training}
Models are trained to convergence on a validation set containing tasks with the same end goals as those in the training data, but with unseen combinations of terrain rewards/penalties. 
The {describer} is trained with typical seq2seq cross-entropy-based supervised learning.
The instructor/executor pair is trained with imitation learning using DAgger~\cite{ross-etal-2011-reduction}. To train the {instructor}, we generate a synthetic instruction for each subtask.
Because the description, which is \textbf{not} shown to the {executor}, conveys terrain rewards/penalties, we train the instructor to decode them as well, e.g. in `\texttt{go} \texttt{to} \texttt{lava} \texttt{and} \texttt{place} \texttt{road.} \ttt{avoid} \texttt{walking} \texttt{on} \texttt{water}.' 
Further details are provided in Appendix \ref{app:training}.

\section{Experiments}
\label{section:experiments}
\noindent\textbf{Demonstration Following}\quad
We test agents 15 times for each evaluation task, using demonstrations in 5 randomly-generated maps each paired with 3 unique maps in which to replicate the task. \\
\noindent\textbf{Description Following}\quad
We use the same task instances as the previous scenario, but provide the ground truth task description directly to the agent.\\
\noindent\textbf{Instruction Following}\quad
To set an upper bound for instructor performance, we evaluate the performance of the executor given {oracle instructions}. 

Our main evaluation metric is the binary completion of the demonstrated task. To measure adherence to terrain constraints, we track the average number of reward/penalty cell traversals and compare to an oracle baseline. 
To measure accuracy against the oracle text, we use exact match computed as a binary sentence-level score. We note that this accuracy does \textbf{not} imply high performance on the benchmark, as the lower-level agents must also understand the text in order to ultimately execute the correct low-level actions to complete the task.

\subsection{Baselines}
\paragraph{Nonverbal Baseline}
To test the effect of computing a latent text representation of the high-level task,
we compare against a nonverbal baseline (\textbf{NV Baseline}) that at each time step computes a continuous representation of the demonstration trajectory instead of encoding a predicted text description. The architecture resembles that of the {executor} module, with a transformer encoder over demonstration transitions (as in the {describer}) rather than text description. Further details are provided in Appendix \ref{app:model:nvb}.
\resizebox{.95\columnwidth}{!}{
  \begin{minipage}{\columnwidth}
\begin{align*}
a_i = f_{\text{exec}}(&M_{1:n}^{\text{dem}}, I_{1:n}^{\text{dem}}, a_{1:n}^\text{dem}, r_{1:n}^\text{dem} ,M_{1:i}, I_{1:i}, a_{1:i-1}) 
\end{align*}
  \end{minipage}
}
\vspace{.5mm}

\paragraph{Latent Language Description Only}
We also compare against a second baseline that conditions the agent's policy on a latent language description (\textbf{LLD}), but does not leverage language at the level of intermediate subtask planning. The LLD architecture resembles the HLLP without the instructor module.
\resizebox{.95\columnwidth}{!}{
\begin{minipage}{\columnwidth}
\begin{align*}
D = f_{\text{descr}}(&M_{1:n}^{\text{dem}}, I_{1:n}^{\text{dem}}, a_{1:n}^\text{dem}, r_{1:n}^\text{dem})\\
a_i = f_{\text{exec}}(& M_{1:i}, I_{1:i}, a_{1:i-1}; D) 
\end{align*}
  \end{minipage}
}
\begin{table*}[t!]
\parbox{0.55\textwidth}{
\vspace{-1.68cm}
    \centering
    \scriptsize
    \setlength{\tabcolsep}{2pt}

\begin{tabular}{lrrrrr|rr}
\toprule
\multirow{3}*{\textbf{EM (\%)}}& &  \multicolumn{4}{c}{\textbf{Describer}} & \multicolumn{2}{c}{\textbf{Instructor}} \\
 & \textbf{\# Eval} & \multicolumn{2}{c}{\textbf{Valid}} & \multicolumn{2}{c}{\textbf{Eval}}  & \multicolumn{2}{c}{\textbf{Eval}}   \\
{} &     \textbf{Tasks} &           \multicolumn{1}{c}{\textbf{Full}} & \multicolumn{1}{c}{\textbf{Goal}} & \multicolumn{1}{c}{\textbf{Full}} & \multicolumn{1}{c}{\textbf{Goal}} & \multicolumn{1}{c}{\textbf{All}} & \multicolumn{1}{c}{\textbf{Last}} \\ \midrule

\textbf{Random Split}      & $\mathbf{15140}$ &                      $\mathbf{84.3}$ &                      $\mathbf{92.4}$ &                     $\mathbf{69.3}$ &                     $\mathbf{75.7}$ &  $\mathbf{77.4 \pm 5.1}$ &  $\mathbf{79.8 \pm 4.3}$ \\
Navigation                          & $700$ &                               $10.1$ &                               $10.6$ &                               $0.9$ &                               $0.9$ &          $60.1 \pm 16.6$ &           $85.1 \pm 1.8$ \\
Crafting                            & $5400$  &                               $98.0$ &                               $98.9$ &                              $87.4$ &                              $88.0$ &           $88.9 \pm 4.4$ &           $83.2 \pm 4.7$ \\
Craft then Nav                      & $ 880$ &                               $88.1$ &                               $99.4$ &                              $84.0$ &                              $88.1$ &           $89.7 \pm 9.6$ &           $97.0 \pm 1.3$ \\
Build on Terrain                 & $ 6040 $&                               $83.0$ &                               $92.9$ &                              $63.8$ &                              $71.7$ &           $78.0 \pm 8.1$ &           $81.7 \pm 5.6$ \\
Cover Terrain                    & $1680$ &                               $71.5$ &                               $98.5$ &                              $59.5$ &                              $84.3$ &           $60.7 \pm 5.1$ &           $52.7 \pm 3.4$ \\
Clear Items                      & $400$ &                               $95.2$ &                               $95.2$ &                              $37.0$ &                              $37.5$ &          $72.2 \pm 10.0$ &          $72.9 \pm 11.0$ \\
\midrule 
\textbf{Hid. Subtask}             &  $\mathbf{8900}$ &                      $\mathbf{84.8}$ &                      $\mathbf{91.4}$ &                     $\mathbf{14.5}$ &                     $\mathbf{15.8}$ &  $\mathbf{43.6 \pm 4.0}$ &  $\mathbf{16.5 \pm 4.8}$ \\
\textbf{Hid. Use Case}            &  $\mathbf{12860}$&                      $\mathbf{84.1}$ &                      $\mathbf{90.3}$ &                     $\mathbf{19.7}$ &                     $\mathbf{22.2}$ &  $\mathbf{40.5 \pm 5.0}$ &  $\mathbf{17.7 \pm 6.8}$ \\
\textbf{Hid. Terr Destn} &$\mathbf{6520}$ &                      $\mathbf{84.9}$ &                      $\mathbf{91.8}$ &                      $\mathbf{0.0}$ &                      $\mathbf{0.0}$ &  $\mathbf{26.5 \pm 2.1}$ &   $\mathbf{5.1 \pm 1.4}$ \\
\textbf{Length Gen.}              &  $\mathbf{5445}$  &                      $\mathbf{85.2}$ &                      $\mathbf{92.0}$ &                     $\mathbf{69.7}$ &                     $\mathbf{92.9}$ &  $\mathbf{62.9 \pm 5.5}$ &  $\mathbf{63.8 \pm 8.1}$ \\
\bottomrule
\end{tabular}
    \caption{Describer and Instructor exact match (EM) against gold references. Describer EM shown for \textbf{Full} text, and first sentence describing end \textbf{Goal}. Validation tasks have same end goals as train, but novel terrain reward/penalty combinations. 
    Instructor EM shown for \textbf{All} and \textbf{Last} instructions given.
    }
    \vspace{-0.5em}
    \label{tab:describer_longform}
}
    \hfill
    \parbox[t][][t]{0.42\textwidth}{
    

\centering
\scriptsize
\setlength{\tabcolsep}{3pt}

\begin{tabular}{lrrr}
\toprule
\textbf{Completion (\%)}  &  \textbf{NV Baseline} &    \textbf{LLD} &   \textbf{HLLP} \\
\midrule \multicolumn{4}{l}{\textbf{Demonstration Following}} \\ \midrule 
\textbf{Overall}         &  $\mathbf{25.2 \pm 7.0}$ &  $\mathbf{65.1 \pm 3.2}$ &  $\mathbf{68.4 \pm 2.2}$ \\
Navigation               &           $45.6 \pm 2.6$ &           $40.5 \pm 1.3$ &           $46.5 \pm 2.9$ \\
Crafting                 &          $44.4 \pm 13.7$ &           $79.6 \pm 3.2$ &           $85.5 \pm 1.7$ \\
Craft then Nav &          $45.4 \pm 14.3$ &           $89.4 \pm 1.8$ &           $95.1 \pm 1.4$ \\
Build on Terrain       &            $9.1 \pm 2.7$ &           $54.4 \pm 4.1$ &           $63.0 \pm 3.4$ \\
Cover Terrain         &            $5.4 \pm 2.9$ &           $61.2 \pm 4.0$ &           $37.9 \pm 1.7$ \\
Clear Items           &           $11.6 \pm 5.6$ &           $39.3 \pm 0.6$ &           $27.0 \pm 6.3$ \\
\midrule \multicolumn{4}{l}{\textbf{Ground Truth Description Following}}  \\  \midrule   
\textbf{Overall}         &                       -- &  $\mathbf{76.7 \pm 3.6}$ &  $\mathbf{82.1 \pm 2.5}$ \\
Navigation               &                       -- &           $93.9 \pm 2.3$ &           $96.2 \pm 2.9$ \\
Crafting                 &                       -- &           $86.0 \pm 3.3$ &           $92.0 \pm 1.8$ \\
Craft then Nav &                       -- &           $90.1 \pm 1.5$ &           $95.9 \pm 1.6$ \\
Build on Terrain       &                       -- &           $67.2 \pm 4.7$ &           $81.3 \pm 4.2$ \\
Cover Terrain         &                       -- &           $64.8 \pm 4.2$ &           $43.8 \pm 2.5$ \\
Clear Items           &                       -- &           $85.8 \pm 3.8$ &           $67.4 \pm 9.1$ \\
\midrule \multicolumn{4}{l}{\textbf{Ground Truth Instruction Following}}  \\  \midrule   
\textbf{Overall}         &                       -- &                       -- &  $\mathbf{97.2 \pm 1.1}$ \\
Navigation               &                       -- &                       -- &           $95.7 \pm 1.5$ \\
Crafting                 &                       -- &                       -- &           $98.1 \pm 0.9$ \\
Craft then Nav &                       -- &                       -- &           $98.5 \pm 0.9$ \\
Build on Terrain       &                       -- &                       -- &           $96.6 \pm 1.4$ \\
Cover Terrain         &                       -- &                       -- &           $97.3 \pm 1.1$ \\
Clear Items           &                       -- &                       -- &           $95.2 \pm 1.8$ \\
\bottomrule
\end{tabular}

   \caption{Completion rates on random task split}
   \vspace{-0.5em}
    \label{tab:rs_combined}

    }
\end{table*}

\section{Results}
We average performance over 5 training seeds. 
\autoref{tab:describer_longform} shows exact match rates for the describer and instructor, measured for the latter at each new instruction.
\autoref{tab:rs_combined} shows completion rate on the random task
split broken down by category, while \autoref{tab:gensplit_combined} shows generalization splits. 

\subsection{Random Split}
Both agents that leverage a predicted task description (HLLP and LLD) outperform the nonverbal baseline on the random unseen task split.
As shown in \autoref{tab:describer_longform}, the describer module exhibits around 70\% exact match accuracy on a set of unseen tasks and 85\% on a set of novel combinations of seen training tasks and terrain constraints.
The describer properly identifies over 
75\% of unseen tasks, which are conveyed by the first sentence of each description. It struggles with navigation and clearing subtasks, which have uniquely short trajectories.
\begin{table}[t!]
\vspace{-2cm}
    \centering
    \scriptsize
\setlength{\tabcolsep}{3pt}

\begin{tabular}{lrrrrrrrrr}
\toprule
\multirow{2}*{\textbf{$|\text{Traversals}|$}} &   & \multicolumn{2}{c}{\textbf{Oracle}}& \multicolumn{2}{c}{\textbf{NVB}} &    \multicolumn{2}{c}{\textbf{LLD}} &   \multicolumn{2}{c}{\textbf{HLLP}}   \\
& \textbf{\# Tasks}& + & \textminus & + & \textminus & + & \textminus & + & \textminus  \\
\midrule
0 Rew 1 Pen    &      5880 &   -- & $7$ &   -- & $30$ &  -- & $12$ &      -- & $19$ \\
0 Rew 2 Pen     &      5595 &  -- & $17$ &  -- & $63$  &  -- & $29$ &      -- & $39$ \\
1 Rew 0 Pen     &      5490 &   $9 $ & -- &   $8 $ & -- & $8 $ & -- &       $7 $ & -- \\
1 Rew 1 Pen     &     11670 &   $9 $ & $7$ &  $7 $ & $32$ &$8 $ & $ 12$ &      $7$ & $20$ \\
2 Rew 0 Pen     &      5430 &  $17$ & -- &  $15 $ & -- &   $15 $ & -- &  $14 $ & -- \\
\bottomrule
\end{tabular}
    \caption{Average traversals on reward (+) or penalty (\textminus)-giving terrains by agents on random split. Tasks are categorized by the number of such terrain types.}
    \vspace{-0.5em}
    \label{tab:terrain_results}
\end{table}
Description following agents achieve high task completion rates given the ground truth task description (\autoref{tab:rs_combined}, middle). The HLLP agent outperforms the {LLD} baseline by greater than 5\%; however, the latter is more effective at covering and clearing subtasks, which require variable numbers of repeated subtasks depending on the random map.
The executor performs nearly perfect given oracle instructions (\autoref{tab:rs_combined}, bottom), indicating
most description following errors are made by the instructor.

\noindent\textbf{Adherence to Terrain Constraints}\quad
\autoref{tab:terrain_results} depicts the rate at which demonstration following agents traverse penalty or reward terrains.\footnote{
Tasks may require traversing a penalty terrain on a randomly generated map.} We compare against an oracle traversal frequency.
This comparison is made difficult by the variability among the times taken by agents to either complete a task or hit the 300-step limit. 
However, the results suggest that the HLLP agent is worse at avoiding penalty terrains than the LLD. All agents are close to oracle performance at traversing reward terrains.
\subsection{Generalization Splits}

\begin{figure}[t!]
    \includegraphics[width=\columnwidth]{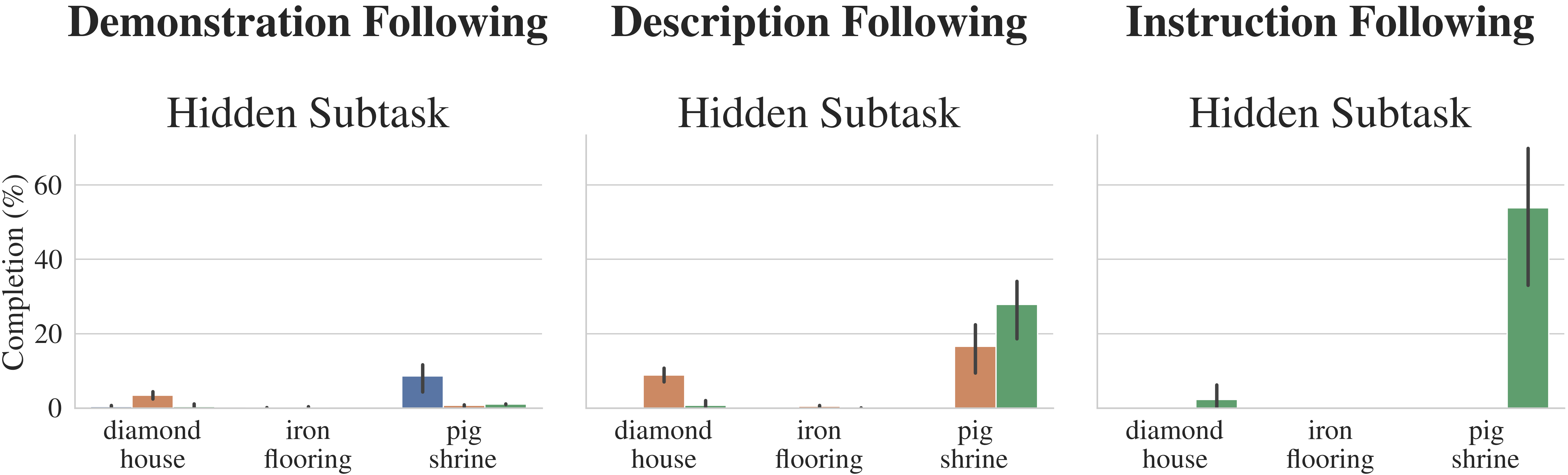}
    \includegraphics[width=\columnwidth]{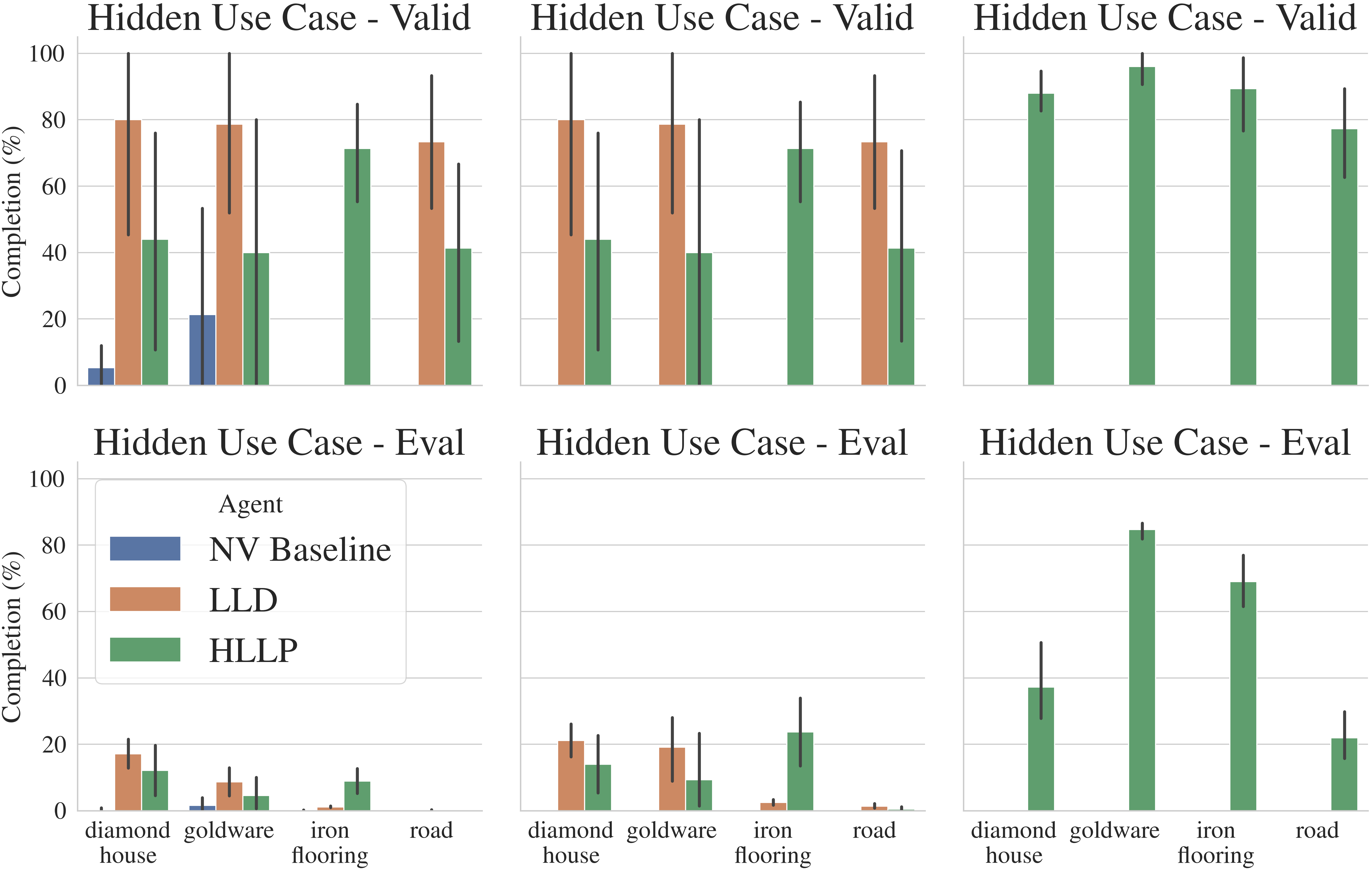}
    \caption{Hidden subtask and use case tests by subtask. }
    \label{fig:gensplit_graphs}
\end{figure}

\noindent\textbf{Hidden Subtask}\quad
Models generally fail to generalize to unseen compositional subtasks. 
The describer identifies only 16\% of the unseen end goals, while the instructor predicts the correct final instruction\footnote{This usually corresponds to the hidden subtask.} at the same rate. 
\autoref{fig:gensplit_graphs} (upper) shows that given gold descriptions, the HLLP agent accomplishes only \texttt{pig} \texttt{shrine} tasks at all, while the LLD also accomplishes \texttt{diamond} {house} at a low rate. 
The executor often fails to handle unseen oracle instructions.\footnote{e.g. the final \ttt{`build} \ttt{diamond} \ttt{house'} instruction.}
We find
that the 
HLLP tends to acquire the correct recipe items, but often does not generate the correct final instruction or perform the right pair of low-level build operations to place the structure. The instructor correctly generates the novel \ttt{pig} \ttt{shrine} concept around 30\% of the time.

\noindent\textbf{Hidden Use Case}\quad
The nonverbal demonstration follower completely fails to generalize tasks to new use cases. The describer module successfully identifies 20\% of unseen use case tasks, but no latent language agent completes more than 5\% from predicted descriptions. 
We observe that completion of the isolated \textit{training} tasks is not perfect (\autoref{fig:gensplit_graphs} middle), indicating that poor performance on this split may be due to a lack of convergence on the subtasks of interest, which underpopulate the training data. The executor module performs well on unseen \ttt{goldware} and \ttt{iron flooring} use cases.

\noindent\textbf{Hidden Terrain Destination}\quad
Agents fail to generalize a terrain observed only as a reward/penalty source to then being a destination for building tasks; particularly for covering tasks. This is the case at all abstraction levels;
the executor given gold instructions completes 55\% of \ttt{build} tasks but only 3\% of \ttt{cover} tasks.
The describer and instructor modules fails to identify the end goal and end instruction at all;
however, in 49\% of describer failure cases, the predicted end goal differs from the ground truth only by the specified destination (e.g. \texttt{on field} instead of the desired \texttt{on water}).


\begin{table}[t!]
    \centering
    \scriptsize
    \setlength{\tabcolsep}{2pt}

\begin{tabular}{lrrr}
\toprule
\textbf{Completion (\%)} & \textbf{NV Baseline} &    \textbf{LLD} &   \textbf{HLLP}  \\
\midrule \multicolumn{4}{l}{\textbf{Demonstration Following}} \\
\midrule
\textbf{Hidden Subtask}             &        $2.5 \pm 1.4$ &   $1.3 \pm 0.4$ &   $0.4 \pm 0.3$ \\
\textbf{Hidden Use Case}            &        $0.3 \pm 0.5$ &   $5.1 \pm 1.5$ &   $5.9 \pm 3.3$ \\
\textbf{Hidden Terr Destn} &        $1.6 \pm 0.9$ &   $4.6 \pm 0.5$ &   $3.7 \pm 0.7$ \\
\textbf{Length Gen.}              &        $6.0 \pm 2.1$ &  $62.6 \pm 3.8$ &  $57.9 \pm 9.0$ \\
\midrule \multicolumn{4}{l}{\textbf{Description Following}} \\    \midrule 
\textbf{Hidden Subtask}             &                   -- &   $7.4 \pm 2.3$ &   $8.0 \pm 3.1$ \\
\textbf{Hidden Use Case}            &                   -- &   $8.2 \pm 1.9$ &  $11.8 \pm 6.9$ \\
\textbf{Hidden Terr Destn} &                   -- &   $1.8 \pm 1.2$ &   $2.8 \pm 1.2$ \\
\textbf{Length Gen.}              &                   -- &  $65.7 \pm 4.1$ &  $60.9 \pm 9.1$ \\
\midrule \multicolumn{4}{l}{\textbf{Instruction Following}} \\    \midrule 
\textbf{Hidden Subtask}             &                   -- &              -- &  $15.6 \pm 7.2$ \\
\textbf{Hidden Use Case}            &                   -- &              -- &  $48.6 \pm 5.0$ \\
\textbf{Hidden Terr Destn} &                   -- &              -- &  $35.3 \pm 7.2$ \\
\textbf{Length Gen.}              &                   -- &              -- &  $96.6 \pm 1.3$ \\
\bottomrule
\end{tabular}
\caption{Completion rates on generalization splits}
    \vspace{-0.5em}
    \label{tab:gensplit_combined}
\end{table}

\noindent\textbf{Length Generalization}\quad
Both latent language agents achieve moderate success on length generalization, particularly relative to the nonverbal baseline (6\% vs 60\%). The describer is extremely successful at identifying long-trajectory tasks.

\subsection{Discussion}
Our results suggest that language serves as an expressive, generalization-promoting representation for one-shot demonstration following agents. 
Our suite of high-level tasks requires an agent to identify task concepts and their roles in composing unique end goal and constraint combinations. Language allows the describer module to communicate such roles succinctly to the other modules, which learn how compositional lexical groundings guide high- and low-level policy decisions in a new context. Learning to encode and plan on the basis of a continuous representation of a demonstration trajectory is otherwise a very challenging task. 
Intermediate-level planning on the basis of LM decoding provides incremental improvements upon nonverbal baselines on a random task split, suggesting improved generalization to other maps and unseen tasks
sampled from the same distribution as those seen during training.  
However, we find that instruction-level latent language does not meaningfully improve \textit{systematic compositional} generalization in either of our evaluation scenarios. 
Reformulating policy search as sequence search simplifies it in certain useful ways--the improved flexibility and interpretability of text-based reasoning allows for pinpointing errors at multiple levels of decision making, 
abstracts away low-level execution decisions that do not pertain to certain forms of generalization, as we observe in our hidden use case results. 
However, a latent language policy alone is not a compositional generalization silver bullet.
Indeed, such challenges remain largely unsolved, though recent approaches have suggested incremental progress in specific cases~\cite{andreas-2020-good, qiu-etal-2021-systematic,conklin-etal-2021-meta}. 
We hope that our benchmark adds to this discourse, and that future work considers our evaluation framework.
We also welcome future work exploring settings with complex subdependencies \textit{under time limits}. To improve stability, our instructor chooses subtasks in an inoptimal canonical order that requires text-based reasoning about high-level tasks, but not spatial reasoning about object proximity. 
\vspace{-1mm}
\section{Conclusion}
Our goal is to design agents that learn new tasks from single examples, with behavior rooted in language.  
This is of broad interest to the NLP community, as one-shot learning of novel tasks combats the typical need to collect and train massive amounts of task data.
This motivated the construction of \env{}, a task environment for testing one-shot learning of complex tasks from demonstrations. \env{} allowed for the development and evaluation of our \textit{hierarchical latent language policy} agent, which performs decision making on the basis of text at multiple levels of abstraction.
We found that models leveraging latent language can improve upon nonverbal alternatives in multiple evaluation scenarios, but that they can struggle with forms of systematic generalization.
We observe that models can accomplish systematically novel tasks provided the correct decision is made at a higher level of abstraction,  which exemplifies how hierarchical latent language provides a mechanism for isolating the level of policy abstraction in which a generalization might occur.

\bibliography{anthology,custom}
\bibliographystyle{acl_natbib}

\clearpage
\appendix

\section*{Contents in Appendices:}
\begin{itemize}
    \item In Appendix~\ref{app:appendix-env}, we provide further details of the \env framework. 
    \item In Appendix~\ref{app:arch}, we describe modeling details of all our proposed agents and baselines.
    \item In Appendix~\ref{app:training}, we provide training and implementation details of our agents. 
    \item In Appendix~\ref{app:additional_results}, we show additional experiment results.
\end{itemize}
\begin{figure*}[t!]
    \centering
    \includegraphics[width=\textwidth]{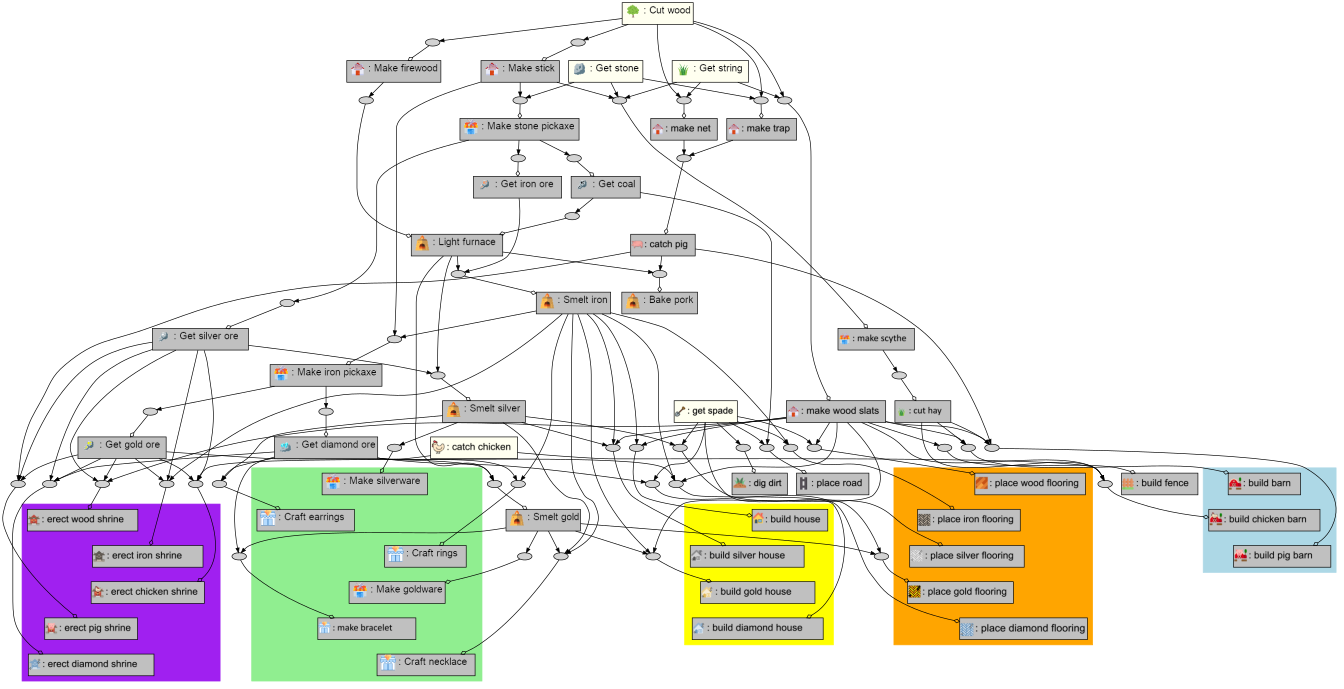}
    \caption{Full subtask dependency graph for the \env{} task environment.}
    \label{fig:full_graph}
\end{figure*}
\section{Environment Details}
\label{app:appendix-env}


As depicted in \autoref{fig:env_overview}, the procedurally generated map (\autoref{fig:env_overview}(a)) is an 8x8 (10x10 with a wall border) grid whose cells may be populated with walls, terrains and interactable objects. Terrains are either \ttt{lava}, \ttt{field} or \ttt{water}. Some objects disappear upon interaction (\ttt{tree}, \ttt{stone}\dots) or transform (\ttt{furnace} $\rightarrow$ \ttt{lit furnace}), or are permanent fixtures (\ttt{lumbershop}, \ttt{workspace} \dots) at which the agent can perform crafting operations. 

The set of possible agent actions comprises directional movement ($\{\texttt{up}, \ttt{down}, \ttt{left}, \ttt{right}\}$, interact actions ($\{\ttt{pick up}, \ttt{use-1}\dots \ttt{use-5}\}$, and place actions ($\{\ttt{place-1}\dots \ttt{place-4}\}$) Subtasks generally have a set of prerequisite subtasks (e.g. \ttt{make stone pickaxe} requires \ttt{get wood} and \ttt{get stone}). The requirements for a subtask do not change across tasks, i.e. \ttt{make stone pickaxe} always requires the same prerequisites and action/location combination.

Crafting tasks require the agent to perform a specific interact action while in the cell of a specific object (\ttt{make stone pickaxe} requires the agent to perform \ttt{use-1} while on top of the \ttt{workspace}. Building tasks require the agent to perform a \ttt{use} action on a cell without an item already inside it. \ttt{place}-based tasks can be performed anywhere regardless of the presence of an item or existing terrain.

If the agent performs actions that render an end goal unattainable (e.g. \ttt{build house on field} but the agent covers all fields with other objects), the game immediately ends and produces a large negative reward. 
\subsection{Task Recipes}
\autoref{fig:full_graph} depicts the full set of \env{} subtasks and their dependencies. 
\subsection{License}
The Mining environment~\cite{sohn-etal-2018-hierarchical} on which our code is licensed under the MIT license.

\begin{table*}[t!]
    \centering
    \scriptsize
    \setlength{\tabcolsep}{2pt}

    \begin{tabular}{llcccccc}
    \toprule
    & & & \multicolumn{5}{c}{\textbf{Action 2}} \\ \cmidrule(lr){4-8}
    \textbf{Base Item} & \textbf{Prerequisites} & \textbf{Action 1} & \ttt{use\_1} & \ttt{use\_2} & \ttt{use\_3} & \ttt{use\_4} & \ttt{use\_5}  \\ \midrule 
    \ttt{flooring}     &  \ttt{spade} & \ttt{place\_2} & \ttt{wood flooring} & \ttt{iron flooring} & \ttt{silver flooring} & \ttt{gold flooring} & \ttt{diamond flooring} \\
    \ttt{barn}     & \ttt{hay, wood slats}& \ttt{use\_2} & \ttt{barn} & & \ttt{chicken barn} & \ttt{pig barn} & \\
    \ttt{house}     &  \ttt{iron, wood slats} & \ttt{use\_3} & \ttt{house} & & \ttt{silver house} & \ttt{gold house} & \ttt{diamond house} \\
    \ttt{shrine}     &  \ttt{gold ore, silver ore} & \ttt{use\_4} & \ttt{wood shrine} & \ttt{iron shrine} & \ttt{chicken shrine} & \ttt{pig shrine} & \ttt{diamond shrine} \\
    \bottomrule
    \end{tabular}
    \caption{List of two-action compositional building/placing recipes}
    \label{tab:compositional_recipes}
\end{table*}
\section{Synthetic Text Generation}
Every subtask in our dependency graph is associated with a single NL phrase, as shown in \autoref{fig:full_graph}. To express task descriptions in NL, we use templates with slots for subtasks, landmarks and terrains; e.g. \texttt{<build\_subtask> on <terrain>.} We take an object-oriented approach to defining end goals, the code for which will be part of our public release. Every goal class, e.g. \texttt{BuildTargetLocationGoal}, \texttt{ClearItemGoal} or \texttt{SubtaskThenDestinationGoal} is associated with a different NL template. Terrain constraints are treated similarly, with slots for terrain type (\texttt{Avoid walking on the <terrain>.} and \texttt{Walking on the <terrain> will reward you}.

To generate oracle instructions during training, we associate with each necessary subtask an instruction to go to a requisite landmark (if necessary) then perform the subtask, e.g. \texttt{go to tree and cut wood} or \texttt{go to water and build house}. If the instructor needs to convey navigational constraints, we append them to the end of the instruction, as in \texttt{go to the workspace and make stone pickaxe. avoid walking on lava}. Figures 8 and 9 show more examples of instructions.

\section{Modeling Details}
\label{app:arch}
In this section, we provide detailed information of our agents.
In Appendix~\ref{app:model:common}, we will describe some common basic components in the agent architecture.
Later on, we will describe each of the proposed agents mentioned in Section~\ref{section:experiments}.

\subsection*{Notations}

We use \textit{game step} $t$ to denote one round of interaction between an agent with the environment.
We use $o_t$ to denote text observation at game step $t$.
$o_t$ may contain different components depending on a specific context, we will describe individual cases in later subsections.
Brackets $[\cdot;\cdot]$ denote vector concatenation.
We use $|s|$ to represent the length of (number of tokens in) a sequence $s$.
We use $h$ and $w$ to denote the height and width of an input image, when the image is flattened, the vector size is $hw$.

\subsection{Common Modules}
\label{app:model:common}

\subsubsection{Text Encoder}
\label{app:model:text_encoder}

We use a transformer-based text encoder, which consists of an embedding layer and a transformer block \citep{vaswani17transformer}.
Specifically, we tokenize an input $o_t$ with the HuggingFace GPT-2 tokenizer\footnote{\url{https://huggingface.co/transformers/model_doc/gpt2.html\#gpt2tokenizer}}.
We convert the tokens into 128-dimension embeddings, the embedding matrix is initialized randomly.

The transformer block consists of a stack of 4 convolutional layers, a self-attention layer, and a 2-layer MLP with a ReLU non-linear activation function in between. 
Within the block, each convolutional layer has 128 filters, with a kernel size of 7.
The self-attention layers use a block hidden size of 128, with 4 attention heads.
Layer normalization \citep{ba16layernorm} is applied after each layer inside the block. 
Following standard transformer training, we add positional embeddings into each block's input.

At every game step $t$, the text encoder encodes $o_t \in \mathbb{R}^{|o_t|}$ and results a representation $h_{o_t} \in \mathbb{R}^{|o_t| \times H}$, $H = 128$ is the hidden size.

\subsubsection{Image Encoder}
\label{app:model:image_encoder}

We propose two image encoder architectures, each tackling a different type of input:

\paragraph{Basic:}
The basic image encoder is adopted from the BabyAI baseline model \citep{chevalier-etal-2018-babyai}.
Specifically, given a symbolic image input $M \in \mathbb{Z}_{\geq0}^{h \times w \times c}$, we use an image bag-of-word (BOW) embedding layer to convert the integer inputs into real-valued embeddings with size $h \times w \times c \times H$,
where $h$, $w$ and $c$ denotes the height, width, and channels of the image, $H = 128$ is the embedding size. 
We sum up the channel dimension, resulting $E_{M} \in \mathbb{R}^{h \times w \times H}$.

Next, the image embeddings are fed into a stacked residual convolutional blocks:
\begin{equation}
\begin{aligned}
    h^{l+1} &= \nresblock^{l}(h^{l}),\\
    h^{0} &= E_{M}.\\
\end{aligned}
\end{equation}
Each residual block consists of two convolutional layers, with kernel size of 3 and output channel size of 128.
Batch normalization is applied after every convolutional layer, followed by a ReLU non-linear activation function. 
Before the last ReLU, we apply a residual connection, which adds the block input into the output of the last batch norm layer.

The output size of the stacked residual blocks is $h \times w \times H$, we flatten its spatial dimensions to result the image encoding $h_M \in \mathbb{R}^{hw \times H}$.

\paragraph{Consecutive:}
In the consecutive image encoder, we aim to capture the difference between two consecutive images. 
Given two images $M_{t-1} \in \mathbb{Z}_{\geq0}^{h \times w \times c}$ and $M_t \in \mathbb{Z}_{\geq0}^{h \times w \times c}$, we first compute their difference $M_{\text{diff}} \in \mathbb{Z}^{h \times w \times c}$. 
We convert the integer inputs into real-valued vectors using image BOW embedding layers, resulting $E_{t-1} \in \mathbb{R}^{h \times w \times H}$, $E_{t} \in \mathbb{R}^{h \times w \times H}$ and $E_{\text{diff}} \in \mathbb{R}^{h \times w \times H}$.
Note $M_{\text{diff}}$ uses a separate image BOW embedding layer. 

To aggregate the three image embeddings, we feed their concatenation into an Multilayer Perceptron (MLP):
\begin{equation}
    E_{M} = \ntanh(\nlinear([E_{t-1};E_{t};E_{\text{diff}}])),
\end{equation}
where $E_{M} \in \mathbb{R}^{h \times w \times H}$.
We use the same convolutional architecture to produce image encoding $h_M \in \mathbb{R}^{hw \times H}$ as in the basic image encoder. 

\subsubsection{Aggregator}
\label{app:model:aggregator}
To aggregate two input encodings $P \in \mathbb{R}^{|P| \times H}$ and $Q \in \mathbb{R}^{|Q| \times H}$, we use the standard multi-head attention mechanism \citep{vaswani17transformer}.
Specifically, we use $P$ as the \textit{query}, $Q$ as the \textit{key} and \textit{value}. 
This results an output $P_Q \in \mathbb{R}^{|P| \times H}$, where at every time step $i \in [0, |P|)$, $P_Q^i$ is the weighted sum of $Q$, the weight is the attention of $P^i$ on $Q$.
We refer readers to \citep{vaswani17transformer} for detailed information.

We apply a residual connection on top of the multi-head attention mechanism in order to maintain the original information contained in $P$. 
Specifically, 
\begin{equation}
    h_{PQ} = \ntanh(\nlinear([P_Q; P])),
\end{equation}
where $h_{PQ} \in \mathbb{R}^{|P| \times H}$.

\subsubsection{Text Decoder}
\label{app:model:text_decoder}

We use a transformer-based text decoder to generate text.
The decoder consists of a word embedding layer, a stacked transformer blocks and a projection layer.

Similar to the text encoder, the embedding layer is initialized with random embedding matrix.
Inside the transformer block, there is one self attention layer, one multi-head attention layer and a 2-layer MLP with ReLU non-linear activation functions in between.
Taking word embedding vectors as input, the self-attention layer first generates a contextual encoding vectors for the words.
These vectors are then fed into the multi-head attention layer, to compute attention with representations produced by the encoder, which contains information from multiple modalities.
The resulting vectors are fed into the 2-layer MLP.
The block hidden size of this transformer is 128.

Subsequently, the output of the stacked transformer blocks is fed into the projection layer, which is a linear transformation with output size same as the vocabulary size.
We follow \citep{press2016using}, tying the input embeddings and this projection layer.
The logits resulted from the projection layer are then normalized by a softmax to generate a probability distribution over all tokens in the GPT-2 vocabulary.

Following common practice, we  use a mask to prevent the decoder transformer to access ``future'' information during training.
We set the max number of generated tokens to be 30.
During inference, the decoder will stop generating whenever generates the \textit{end-of-sequence} special token, or exhausts all its budget.

\subsection{Hierarchical Latent Language Policy Agent (HLLP)}
\label{app:modeo:hllp}

\subsubsection{Describer}
\label{app:model:describer}
As briefly mentioned in Section~\ref{section:hllp:describer}, the describer module ``summarizes'' a demonstration into a short text, where a demonstration typically a sequence of multi-modal transitions.
As shown in Figure~\ref{fig:agent_2_arch}, at every step $t$ of a demonstration, the transition contains the symbolic images at previous step and current step:  $M_{t-1}$ and $M_t$, and the text input $o_t = [a_{t-1}; r_{t-1}; I_t]$, where $a_{t-1}$, $r_{t-1}$, $I_t$ denote the action taken at previous step, the resulting reward, and the inventory state at current step, respectively. 

We first encode the text input with an text encoder described in Appendix~\ref{app:model:text_encoder}, similarly, we encode the image inputs with an consecutive image encoder described in Appendix~\ref{app:model:image_encoder}.
We subsequently use two attention blocks described in Appendix~\ref{app:model:aggregator} to compute the image encoding's attention over text (tokens), and vice versa, the text encoding's attention over image (pixels).
We average both the attention-aggregated outputs, resulting $h_{\text{img} \rightarrow \text{text}} \in \mathbb{R}^{\times H}$ and $h_{\text{text} \rightarrow \text{img}} \in \mathbb{R}^{\times H}$, to compute the overall representation of this time step:
\begin{equation}
    h_{t} = \ntanh(\nlinear([h_{\text{img} \rightarrow \text{text}}; h_{\text{text} \rightarrow \text{img}}])),
\end{equation}
where $h_{t} \in \mathbb{R}^{\times H}$, $H=128$ is the hidden size.

At the episode level, we use a Transformer-based encoder, with similar architecture to the one in our text encoder.
Specifically, the episode encoder is a stacked 2-layer Transformer blocks, which outputs $h_{\text{demo}^i} \in \mathbb{R}^{|\text{demo}^i|\times H}$, $|\text{demo}^i|$ is the number of steps of a demonstration $\text{demo}^i$, $H$ is hidden size.

Finally, we use a text decoder, as described in Appendix~\ref{app:model:text_decoder}, to generate text descriptions.

In the describer module, we use a 2-layer text encoder, a 5-layer image encoder, a 2-layer episode encoder, and a 3-layer decoder.

\subsubsection{Instructor}
\label{app:model:instructor}

As shown in Figure~\ref{fig:agent_2_arch}, the instructor consists a text encoder, a basic graph encoder, an attention mechanism, a text decoder, and a new instruction classifier. 

Specifically, at a game step $t$, the image encoder takes the image input $M_t$ as input, generates image representations $v_t \in \mathbb{R}^{hw \times H}$, where $h$ and $w$ are the height and width of the image. 
At the same time, the text encoder encodes the text input $o_t = [D; I_t; \text{Instr}_{t-1}]$, where $D$, $I_t$ and $\text{Instr}_{t-1}$ denote the task description (either generated by the describer, or provided by an oracle), the inventory state at current step, and the instruction at previous game step.
The text encoder outputs $w_t \in \mathbb{R}^{|o_t| \times H}$.
Next, an attention block as described in Appendix~\ref{app:model:aggregator} aggregates $v_t$ and $w_t$, resulting $s_t \in \mathbb{R}^{|o_t| \times H}$ that contains information from both modalities, where $|o_t|$ denotes number of tokens in $o_t$.

The new instruction classifier is an MLP switch module that decides whether or not the instruction generated at previous step is still valid (i.e., is it necessary to generate a new instruction):
\begin{equation}
\begin{aligned}
    s'_t &= \nmaskedmean(s_t),\\
    p_t &= \nargmax(L^1(\ntanh(L^0(s'_t)))).\\
\end{aligned}
\end{equation}
In which, $L^0$ and $L^1$ are linear transformations with hidden size of 128 and 2, respectively. 
The output $p_t \in \{0, 1\}$ is the discrete switch.

In the case where $p_t = 0$, we directly pass the instruction generated at previous step along as output; otherwise, a text decoder as described in Appendix~\ref{app:model:text_decoder} will generate a new instruction word-by-word conditioned on $s_t$.

In the describer module, we use a single layer text encoder, a 2-layer image encoder, and a 2-layer decoder.
The text encoder and image encoder are tied with the corresponding layers in the executor module.
During training, we do not update the image encoder.

\subsubsection{Executor}
\label{app:model:executor}

Given the intermediate level text instruction, our executor module translates them into low level actions to interact with the environment. 
As shown in Figure~\ref{fig:agent_2_arch}, the executor consists a text encoder, a basic graph encoder, an attention block, and a recurrent action generator. 

Similar to the instructor module, the image encoder and text encoder convert image input ($M_t$) and text input ($I_t$ and $\text{Instr}_{t}]$) into hidden representations.
Note in the executor, to facilitate interaction between the instruction $\text{Instr}_{t}]$ with other text inputs, we encode $I_t$ and $\text{Instr}_{t}]$ separately and aggregate them using an attention mechanism. 

Subsequently, given the image representation $v_t$ and the aggregated text representation $w_t$, we apply attention block (as described in Appendix~\ref{app:model:aggregator}) from both directions:
\begin{equation}
\begin{aligned}
    h_{vw} &= \nattention(v_t, w_t), \\
    h_{wv} &= \nattention(w_t, v_t), \\
    h'_{vw} &= \nmaskedmean(h_{vw}), \\
    h'_{wv} &= \nmaskedmean(h_{wv}), \\
    s_t &= \ntanh(\nlinear([h'_{vw}; h'_{wv}])),\\
\end{aligned}
\end{equation}
in which, $s_t \in \mathbb{R}^{H}$, $H = 128$ is hidden dimension.

In order to encourage the action generator to condition on history information, we equip it with a recurrent memory \citep{cho2014gru}:
\begin{equation}
    s_{1:t} = \ngru(s_t, s_{1:t-1}), \\
\end{equation}
the hidden size of the GRU is 128. 
We stack an MLP on top of the recurrent memory to obtain the output distribution over all actions:
\begin{equation}
\begin{aligned}
    h_t &= \ntanh(\nlinear(s_{1:t})), \\
    p_{a_t} &= \nsoftmax(\nlinear(h_t)),\\
    a_t &= \nargmax(p_{a_t}).\\
\end{aligned}
\end{equation}

In the executor module, we use a single layer text encoder and a 2-layer image encoder.
The text encoder and image encoder are tied with the corresponding layers in the instructor module.
During training, we do not update the text encoder.

\subsection{Latent Language Description Only Baseline (LLD)}
\label{app:model:lld}

The LLD baseline shares the same describer architecture, and a similar executor architecture with HLLP, its main difference is the absence of an instructor. 

In its executor, at a game step $t$, the inputs are an image $M_t$ and a short text $o_t = [D; I_t]$, where $D$ is the description generated by the describer (or the oracle description during training), $I_t$ is the agent's inventory state.
To obtain the text representation $w_t$, the LLD agent simply encode $o_t$ with the text encoder as described in Appendix~\ref{app:model:text_encoder}, without performing attention between $D$ and $I_t$ (as in HLLP).
The rest of the executor components are identical to HLLP (Appendix~\ref{app:model:executor}).

In the LLD baseline, we use a single layer text encoder and a 2-layer image encoder.

\subsection{Nonverbal Baseline (NV)}
\label{app:model:nvb}
In the nonverbal baseline, we do not use language as latent representations between modules. 
Specifically, given a demonstration $\text{demo}^i$, we use a describer similar to the one outlined in Appendix~\ref{app:model:describer}, but without decoding the demonstration representation into text.
The output of the describer is $h_{\text{demo}^i} \in \mathbb{R}^{|\text{demo}^i|\times H}$, where $|\text{demo}^i|$ is the number of steps in $\text{demo}^i$, $H$ is hidden size.

In our nonverbal baseline's executor, at game step $t$, a text encoder encodes the inventory state $I_t$ into $w_t$; an image encoder encodes an input image $M_t$ into $v_t$.
We use multi-head attention blocks (Appendix~\ref{app:model:aggregator}) to aggregate information carried by image ($v_t$), text ($w_t$), and demonstration representation ($h_{\text{demo}^i}$):
\begin{equation}
\begin{aligned}
    h'_{\text{demo}^i} &= \nmaskedmean(h_{\text{demo}^i}), \\
    h_{\text{demo} \rightarrow \text{img}} &= \nattention(h'_{\text{demo}^i}, v_t), \\
    h_{\text{demo} \rightarrow \text{text}} &= \nattention(h'_{\text{demo}^i}, w_t), \\
    h_{\text{text} \rightarrow \text{img}} &= \nattention(w_t, v_t), \\
    h_{\text{img} \rightarrow \text{text}} &= \nattention(v_t, w_t), \\
    h'_{\text{text} \rightarrow \text{img}} &= \nmaskedmean(h_{\text{text} \rightarrow \text{img}}), \\
    h'_{\text{img} \rightarrow \text{text}} &= \nmaskedmean(h_{\text{img} \rightarrow \text{text}}). \\
\end{aligned}
\end{equation}
Subsequently, we use an MLP to combine them:
\begin{equation}
\begin{aligned}
    h_{\text{combined}} = &[h'_{\text{demo}^i}; \\
    &h_{\text{demo} \rightarrow \text{img}}; h_{\text{demo} \rightarrow \text{text}};\\
    &h'_{\text{text} \rightarrow \text{img}} ; h'_{\text{img} \rightarrow \text{text}}], \\
    s_t = &\ntanh(\nlinear(h_{\text{combined}})),\\
\end{aligned}
\end{equation}
in which, the output $s_t \in \mathbb{R}^{H}$, $H = 128$ is hidden dimension.

The remainder of the executor is identical to the executor used in the HLLP agent, as described in Appendix~\ref{app:model:executor}.

In the nonverbal baseline, we use a single layer text encoder and a 2-layer image encoder.

\section{Training and Implementation Details}
\label{app:training}

For all experiments, we use \emph{Adam} \citep{kingma14adam} as the optimizer.
The learning rate is set to 0.001 with a clip gradient norm of 5.

\subsection{Describer Training via Supervised Learning}

We use a set of pre-collected expert demonstrations paired with ground-truth descriptions to train the describer module in HLLP. 
Because demonstrations are long sequences of agent transitions, which can be memory consuming, we cut long demonstrations and only keep their last 100 transition steps. 
Since the length of demonstration varies, we speed up training by sorting the data points by their demonstration length, and split them by buckets with a bucket size of 2,000.
For every mini-batch (we use a batch size of 20), we first randomly sample a bucket, then randomly sample a batch of data point from that bucket. 
We train the describer for 5 million episodes (250,000 batches).

\subsection{Description Follower Training via DAgger}

We train the description follower modules (instructor and executor in HLLP, executor in LLD, and the entire nonverbal baseline) using DAgger \citep{ross-etal-2011-reduction}, an imitation learning method. 

Specifically, during the training process, the agent starts with totally following the expert demonstrations, then we gradually let the agent to take over the control. The expert takes the form of a greedy oracle that identifies eligible and necessary subtask landmarks, navigates to them according to a traversal cost graph that considers terrain rewards and penalties, then performs the subtask. We initially had the oracle complete whichever eligible subtask required the fewest steps. However, this led to training instability due to the compounded difficulty of inferring required subtasks and selecting an eligibility-adherent completion order based on distances in a random map. Instead, we choose the first eligible subtask in a canonically-ordered list.

We collect such trajectories (i.e., sequences of transitions, along the expert demonstrations if the agent takes over control), without updating the network, into a replay buffer of size 500,000.
We periodically (after every 5 data collection steps) sample batches of transitions from the replay buffer, and update the network.
Specifically, following the training strategy used in the recurrent DQN literature~\citep{hausknecht2015deep,yuan2018counting}, we sample batches of transition sequences (of length 8), we use the first 4 transitions to estimate the recurrent states, and the last 4 transitions for updating the model parameters.
We use a mini-batch of size 32 in replay data collection, and a batch size of 64 for update.
We linearly anneal the fraction of expert assistance in DAgger from 100\% to 1\% within 500,000 episodes.

When training the HLLP agent, as depicted in Figure~\ref{fig:agent_2_arch}, we tie the encoder parameters between the instructor and the executor. 
In which, the image encoder is only updated through the executor loss, whereas the text encoder is only updated through the instructor loss. 
To stabilize the training, we update the instructor and executor modules in an alternate manner, with a frequency of 2,000 (experience data collection) episodes. 

We train the description following agents for 1 million episodes maximally, however, in practice, the agents mostly converge sooner. 
We set an patience of 100,000 episodes, the training process will terminate if there is no improvement within this period.

\subsection{Resources}
We use a mixture of Nvidia V100/P100/P40 GPUs to train all models; on average experiments (training with environment simulation) take 3-4 days, but the wall clock time can vary.

\section{Supplementary Results}
\label{app:additional_results}
\begin{table}[t!]
    \centering
    \scriptsize
    \setlength{\tabcolsep}{2pt}
    \begin{tabular}{lrrrr}
\toprule
{} & \multicolumn{2}{c}{\textbf{Valid}} & \multicolumn{2}{c}{\textbf{Eval}} \\
{} & \textbf{Full Task} & \textbf{End Goal} & \textbf{Full Task} & \textbf{End Goal} \\
\midrule
\midrule \textbf{Random Split}      &    $\mathbf{84.3}$ &    $\mathbf{92.4}$ &    $\mathbf{69.3}$ &    $\mathbf{75.7}$ \\ 
Navigation                          &             $10.1$ &             $10.6$ &              $0.9$ &              $0.9$ \\
Crafting                            &             $98.0$ &             $98.9$ &             $87.4$ &             $88.0$ \\
Craft then Nav                      &             $88.1$ &             $99.4$ &             $84.0$ &             $88.1$ \\
Building on Terrain                 &             $83.0$ &             $92.9$ &             $63.8$ &             $71.7$ \\
Covering Terrain                    &             $71.5$ &             $98.5$ &             $59.5$ &             $84.3$ \\
Clearing Items                      &             $95.2$ &             $95.2$ &             $37.0$ &             $37.5$ \\
\midrule \textbf{Hidden Subtask}             &    $\mathbf{84.8}$ &    $\mathbf{91.4}$ &    $\mathbf{14.5}$ &    $\mathbf{15.8}$ \\ 
Crafting                            &             $97.8$ &             $98.4$ &             $36.1$ &             $36.4$ \\
Craft then Nav                      &             $88.2$ &             $98.3$ &             $32.8$ &             $32.8$ \\
Building on Terrain                 &             $84.6$ &             $93.0$ &              $6.4$ &              $7.5$ \\
Covering Terrain                    &             $74.9$ &             $97.6$ &              $7.2$ &             $12.1$ \\
 \midrule \textbf{Hidden Use Case}            &    $\mathbf{84.1}$ &    $\mathbf{90.3}$ &    $\mathbf{19.7}$ &    $\mathbf{22.2}$ \\
Crafting                            &             $95.1$ &             $95.6$ &             $29.1$ &             $29.3$ \\
Craft then Nav                      &             $90.4$ &             $99.7$ &             $46.2$ &             $47.5$ \\
Building on Terrain                 &             $84.6$ &             $93.9$ &             $20.3$ &             $23.5$ \\
Covering Terrain                    &             $75.3$ &             $97.7$ &              $4.0$ &              $7.4$ \\
 \midrule\textbf{Hidden Terrain Destination} &    $\mathbf{84.9}$ &    $\mathbf{91.8}$ &     $\mathbf{0.0}$ &     $\mathbf{0.0}$ \\
Building on Terrain                 &             $84.0$ &             $94.4$ &              $0.0$ &              $0.0$ \\
Covering Terrain                    &             $71.9$ &             $97.7$ &              $0.0$ &              $0.0$ \\
\midrule  \textbf{Hidden Length}              &    $\mathbf{85.2}$ &    $\mathbf{92.0}$ &    $\mathbf{69.7}$ &    $\mathbf{92.9}$ \\ 
Crafting                            &             $97.3$ &             $98.1$ &             $95.6$ &             $99.1$ \\
Craft then Nav                      &             $89.9$ &             $99.6$ &             $89.1$ &            $100.0$ \\
Building on Terrain                 &             $82.9$ &             $93.2$ &             $74.4$ &             $91.0$ \\
Covering Terrain                    &             $76.8$ &             $97.1$ &             $58.9$ &             $92.6$ \\
Clearing Items                      &             $98.8$ &             $99.1$ &            $100.0$ &            $100.0$ \\
\bottomrule
\end{tabular}

    \caption{Expanded performance of Describer module against gold references in all splits and task categories. Validation scores for task categories not in an eval set are not shown.}
    \label{tab:describer_long}
\end{table}
\begin{table}[t!]
    \centering
    \scriptsize
    \setlength{\tabcolsep}{2pt}
\begin{tabular}{lrrrr}
\toprule
 & \textbf{NV Baseline} &   \textbf{LLD} &   \textbf{HLLP} \\
\midrule
\midrule \multicolumn{4}{l}{\textbf{Demonstration Following}} \\    \midrule 
\textbf{Overall}    &        $1.6 \pm 0.9$ &  $4.6 \pm 0.5$ &  $3.7 \pm 0.7$ \\
Building on Terrain &        $2.5 \pm 1.5$ &  $7.4 \pm 0.8$ &  $6.0 \pm 1.1$ \\
Covering Terrain               &        $0.0 \pm 0.0$ &  $0.1 \pm 0.0$  &    $0.0 \pm 0.0$ \\
\midrule \multicolumn{4}{l}{\textbf{Ground Truth Description Following}} \\    \midrule 
\textbf{Overall}    &                   -- &  $1.8 \pm 1.2$ &    $2.8 \pm 1.2$ \\
Building on Terrain &                   -- &  $2.9 \pm 2.0$ &    $4.5 \pm 1.9$ \\
Covering Terrain               &                   -- &  $0.0 \pm 0.0$  &    $0.1 \pm 0.1$ \\
\midrule \multicolumn{4}{l}{\textbf{Ground Truth Instruction Following}} \\    \midrule 
\textbf{Overall}    & -- &  -- &   $35.3 \pm 7.2$ \\
Building on Terrain & -- &  -- &  $55.1 \pm 11.2$ \\
Covering Terrain    & -- &  --  &    $3.1 \pm 0.8$ \\
\bottomrule
\end{tabular}
    \caption{Performance on hidden terrain destination split broken down by task category}
    \label{app:hidden_terrain_dest}
\end{table}
\begin{table}[t!]
    \centering
    \scriptsize
    \setlength{\tabcolsep}{2pt}
    \begin{tabular}{lrrrrr}
\toprule
 & \textbf{\# Tasks} & \textbf{NVB} &    \textbf{LLD} &    \textbf{HLLP} \\
\midrule
\midrule \multicolumn{4}{l}{\textbf{Demonstration Following}} \\    \midrule 
\textbf{Overall}   &  &        $6.0 \pm 2.1$ &  $62.6 \pm 3.8$ &   $57.9 \pm 9.0$ \\
Crafting         & 1905   &       $29.9 \pm 8.1$ &  $82.5 \pm 3.5$ &  $86.0 \pm 11.6$ \\
Build on Terr &   6330 &     $4.9 \pm 2.9$ &  $58.9 \pm 4.5$ &  $69.6 \pm 13.2$ \\
Cover Terr      & 7830         &        $0.3 \pm 0.4$ &  $59.7 \pm 3.9$ &   $41.1 \pm 5.4$ \\
Craft then Nav       & 165    &       $36.4 \pm 3.8$ &  $91.8 \pm 4.6$ &   $88.6 \pm 8.9$ \\
Clear Itm   & 105   &       $18.5 \pm 9.1$ &  $87.8 \pm 5.1$ &  $42.1 \pm 11.3$ \\
\midrule \multicolumn{4}{l}{\textbf{Ground Truth Description Following}} \\    \midrule 
\textbf{Overall}    &        &           -- &  $65.7 \pm 4.1$ &   $60.9 \pm 9.1$ \\
Crafting            &       1905&            -- &  $82.8 \pm 3.4$ &  $86.3 \pm 11.6$ \\
Build on Terr &   6330 &                   -- &  $62.4 \pm 4.9$ &  $75.1 \pm 13.8$ \\
Cover Terr       & 7830            &                   -- &  $63.3 \pm 4.1$ &   $42.9 \pm 5.3$ \\
Craft then Nav      & 165       &                   -- &  $91.8 \pm 4.6$ &   $88.4 \pm 9.2$ \\
Clear Itm    & 105    &                   -- &  $87.8 \pm 5.1$ &  $42.1 \pm 11.3$ \\
\midrule \multicolumn{4}{l}{\textbf{Ground Truth Instruction Following}} \\    \midrule 
\textbf{Overall}    & &                   -- &              -- &   $96.6 \pm 1.3$ \\
Crafting          &       1905  &                   -- &              -- &   $97.4 \pm 1.9$ \\
Build on Terr &   6330 &                   -- &              -- &      $97.1 \pm 1.2$ \\
Cover Terr         & 7830         &                   -- &              --  &   $95.9 \pm 1.4$ \\
Craft then Nav      & 165       &                   -- &              -- &   $98.7 \pm 0.9$ \\
Clear Itm     & 105   &                   -- &              -- &   $96.6 \pm 1.6$ \\
\bottomrule
\end{tabular}
    \caption{Length generalization results}
    \label{app:length_gen}
\end{table}
\autoref{tab:describer_longform} Shows describer module exact match performance against gold references in all splits and task categories.

\autoref{app:hidden_terrain_dest} shows full task completion performance by agents on the hidden terrain destination generalization set set decomposed by task category.
\autoref{app:length_gen} shows the same for the length generalization set.

\autoref{fig:oracle_unrolls} depicts example unrolled trajectories produced by the oracle. \autoref{fig:gensplit_unrolls} depicts example failure cases by the HLLP agent on the generalization splits.
\begin{figure*}[]
    \centering
    \scriptsize
    \setlength{\parindent}{0pt} 

\begin{tabular}{V{\columnwidth}|V{\columnwidth}}
\toprule
\begin{verbatim}
build fence on silver flooring, then reach the jeweler. 
avoid walking on the field. walking on the lava will reward you.
================================================
I0: cut wood, stepping on lava and avoiding field (9 steps)
I1: get stone, stepping on the lava and avoiding the field 
    (3 steps)
I2: get string, stepping on the lava and avoiding the field
    (4 steps)
I3: get spade, stepping on the lava and avoiding the field
    (4 steps)
I4: make stick, stepping on the lava and avoiding the field
    (6 steps)
I5: make wood slats (1 steps)
I6: make stone pickaxe, stepping on the lava and avoiding
    the field (7 steps)
I7: get coal, stepping on the lava and avoiding the field 
    (4 steps)
I8: get silver ore, stepping on the lava and avoiding the 
    field (11 steps)
I9: light furnace, stepping on the lava and avoiding the 
    field (3 steps)
I10: smelt silver (1 steps)
I11: place silver flooring on empty cell, stepping on the lava 
    and avoiding the field (3 steps)
I12: build fence on silver flooring (1 steps)
I13: go to jeweler, stepping on the lava and avoiding the 
    field (5 steps)
game ended after 62 steps
\end{verbatim}&
\begin{verbatim}
make net and place silver flooring covering all the 
water in any order. avoid walking on the field.
================================================
I0: cut wood, avoiding the field (5 steps)
I1: get stone, avoiding the field (7 steps)
I2: get string, avoiding the field (7 steps)
I3: get spade, avoiding the field (7 steps)
I4: make firewood, avoiding the field (6 steps)
I5: make stick (1 steps)
I6: make net (1 steps)
I7: make stone pickaxe, avoiding the field (5 steps)
I8: get silver ore, avoiding the field (2 steps)
I9: light furnace, avoiding the field (10 steps)
I10: smelt silver (1 steps)
I11: place silver flooring covering water, avoiding the field 
    (4 steps)
I12: place silver flooring covering water, avoiding the field 
    (3 steps)
I13: place silver flooring covering water, avoiding the field 
    (3 steps)
I14: place silver flooring covering water, avoiding the field 
    (3 steps)
I15: place silver flooring covering water, avoiding the field 
    (3 steps)
I16: place silver flooring covering water, avoiding the field 
    (3 steps)
I17: place silver flooring covering water, avoiding the field 
    (3 steps)
game ended after 88 steps
\end{verbatim}\\ \midrule
\begin{verbatim}
dig dirt covering all the water, then reach the workspace.
================================================
I0: get spade (8 steps)
I1: dig dirt covering water (2 steps)
I2: dig dirt covering water (2 steps)
I3: dig dirt covering water (3 steps)
I4: dig dirt covering water (2 steps)
I5: dig dirt covering water (2 steps)
I6: dig dirt covering water (2 steps)
I7: dig dirt covering water (3 steps)
I8: dig dirt covering water (2 steps)
I9: dig dirt covering water (3 steps)
I10: dig dirt covering water (2 steps)
I11: dig dirt covering water (2 steps)
game ended after 32 steps
\end{verbatim} & 
\begin{verbatim}
clear all of the grasses and the irons.
================================================
I0: cut wood (6 steps)
I1: get stone (5 steps)
I2: get string (5 steps)
I3: make stick (12 steps)
I4: make stone pickaxe (2 steps)
I5: make scythe (1 steps)
I6: get iron ore (4 steps)
I7: get iron ore (3 steps)
I8: cut hay (4 steps)
I9: cut hay (4 steps)
I10: cut hay (10 steps)
game ended after 56 steps 
\end{verbatim}
\\
\midrule 
\begin{verbatim}
 build pig barn on dirt and build diamond house on silver flooring 
 in any order. 
 ================================================
I0: cut wood (8 steps)
I1: get stone (3 steps)
I2: get string (2 steps)
I3: get spade (12 steps)
I4: make stick (12 steps)
I5: make trap (1 steps)
I6: make net (1 steps)
I7: make wood slats (1 steps)
I8: make stone pickaxe (7 steps)
I9: catch pig (3 steps)
I10: make scythe (3 steps)
I11: get coal (16 steps)
I12: get iron ore (15 steps)
I13: get silver ore (5 steps)
I14: cut hay (5 steps)
I15: dig dirt on empty cell (2 steps)
I16: light furnace (12 steps)
I17: build pig barn on dirt (13 steps)
I18: smelt iron (12 steps)
I19: smelt silver (1 steps)
I20: make iron pickaxe (4 steps)
I21: get diamond ore (3 steps)
I22: place silver flooring on empty cell (5 steps)
I23: build diamond house on silver flooring (2 steps)
game ended after 148 steps (task was completed)
\end{verbatim}
& 
\begin{verbatim}
place diamond flooring on field, then reach the lumbershop. 
================================================
I0: cut wood (11 steps)
I1: get stone (5 steps)
I2: get spade (4 steps)
I3: make stick (6 steps)
I4: make stone pickaxe (7 steps)
I5: get coal (5 steps)
I6: get iron ore (7 steps)
I7: light furnace (6 steps)
I8: smelt iron (1 steps)
I9: make iron pickaxe (6 steps)
I10: get diamond ore (3 steps)
I11: place diamond flooring on field (5 steps)
I12: go to lumbershop (4 steps)
game ended after 70 steps  
\end{verbatim}
\\
\bottomrule
\end{tabular}
\caption{Example unrolled oracle trajectories}
\label{fig:oracle_unrolls}
\end{figure*}
\begin{figure*}[]
    \centering
    \scriptsize
    \setlength{\parindent}{0pt} 

\begin{tabular}{V{\columnwidth}|V{\columnwidth}}
\toprule 
\multicolumn{2}{l}{\textbf{Hidden Subtask}} \\ \midrule 
\begin{verbatim}
 erect pig shrine.
 ================================================
I.0: cut wood
I.1: get stone
I.2: get string
I.3: make stick
I.4: make trap
I.5: make net
I.6: make stone pickaxe
I.7: catch pig
I.8: get coal
I.9: get iron ore
I.10: get silver ore
I.11: light furnace
I.12: smelt iron
I.13: make iron pickaxe
I.14: get gold ore <pig shrine now eligible>
I.15: erect pig shrine <agent erects iron shrine>
I.16: erect pig shrine <agent erects iron shrine on same cell>
<repeats until time limit>
game ended after 300 steps (task incomplete)
\end{verbatim}&
\begin{verbatim}
build diamond house.
================================================
I.0: cut wood
I.1: get stone
I.2: get string
I.3: make stick
I.4: make wood slats
I.5: make stone pickaxe
I.6: get coal
I.7: get iron ore
I.8: light furnace
I.9: smelt iron
I.10: make iron pickaxe
I.11: get gold ore
I.12: get diamond ore <diamond house now eligible>
I.13: erect diamond shrine <agent erects diamond shrine unsuccessfully> 
I.13: erect diamond shrine <agent erects diamond shrine unsuccessfully>
<repeats until time limit>
game ended after 300 steps (task incomplete)
\end{verbatim}\\
\midrule \multicolumn{2}{l}{\textbf{Hidden Use Case}} \\ \midrule 
\begin{verbatim}
place iron flooring covering all the lava and erect pig shrine 
on silver flooring in any order.
================================================
I.0: cut wood 
I.1: get stone 
I.2: get string 
I.3: get spade 
I.4: make stick 
I.5: make trap
I.6: make net
I.7: make stone pickaxe 
I.8: catch pig 
I.9: get coal 
I.10: get iron ore 
I.11: get silver ore 
I.12: light furnace 
I.13: smelt iron
I.14: smelt silver
I.15: make iron pickaxe 
I.16: place iron flooring covering lava 
I.17: place iron flooring covering lava
I.18: place iron flooring covering lava
I.19: place iron flooring covering lava <lava fully covered>
I.20: place iron flooring covering lava 
<repeats until time limit>
game ended after 300 steps (task incomplete, no pig shrine)
\end{verbatim}
&
\begin{verbatim}
build chicken barn on road and get gold ore in any order.
================================================
I.0: cut wood
I.1: get stone
I.2: get string
I.3: catch chicken
I.4: make stick
I.5: make wood slats
I.6: make stone pickaxe
I.7: make scythe
I.8: get coal
I.9: get iron ore
I.10: cut hay
I.11: light furnace
I.12: build chicken barn on empty cell
I.13: smelt iron
I.14: make iron pickaxe
I.15: get gold ore
<repeats until time limit>
game ended after 300 steps (task incomplete, barn not in road)
\end{verbatim} \\
\midrule \multicolumn{2}{l}{\textbf{Hidden Terrain Destination}} \\ \midrule 
\begin{verbatim}
place silver flooring covering all the water.
================================================
I.0: cut wood
I.1: get stone
I.2: get spade
I.3: make stick
I.4: make stone pickaxe
I.5: get coal
I.6: get silver ore
I.7: light furnace
I.8: smelt silver
I.9: place silver flooring covering field 
<repeats until time limit>
game ended after 300 steps (task incomplete, water not covered)
\end{verbatim}
&
\begin{verbatim}
build fence on water.
================================================
I.0: cut wood
I.1: get string
I.2: make wood slats
I.3: build fence on empty cell
<repeats until time limit>
game ended after 300 steps (task incomplete, fence not on water)
\end{verbatim} \\
\bottomrule
\end{tabular}
\caption{Example agent failure cases on generalization splits}
\label{fig:gensplit_unrolls}
\end{figure*}
\end{document}